\documentclass[]{ceurart}

\usepackage{xcolor}
\usepackage{stmaryrd}
\usepackage{amssymb}
\usepackage{underscore}
\usepackage{float}

\providecommand{\doi}[1]{\href{https://doi.org/#1}{#1}}

\begin{document}

\copyrightyear{2026}
\copyrightclause{Copyright for this paper by its authors.
  Use permitted under Creative Commons License Attribution 4.0
  International (CC BY 4.0).}

\conference{ARQNL 2026: Automated Reasoning in Quantified
  Non-Classical Logics, 24 July 2026, Lisbon, Portugal}

\title{First-Order Modal Logic in HOL: Deep and Shallow Embeddings with Automated Faithfulness (Extended Preprint)}

\author[1,2]{Christoph Benzm\"uller}[%
  orcid=0000-0002-3392-3093,
  email=christoph.benzmueller@uni-bamberg.de]
\cormark[1]
\address[1]{University of Bamberg, Bamberg, Germany}
\address[2]{Freie Universit\"at Berlin, Berlin, Germany}

\author[1]{Daniel Kirchner}[%
  orcid=0000-0001-9229-1148,
  email=daniel.kirchner@uni-bamberg.de]

\cortext[1]{Corresponding author.}

\begin{abstract}
We extend, in Isabelle/HOL, the deep-and-shallow embedding
methodology of our prior work from propositional to first-order
modal logic (FML) with constant-domain Kripke semantics. Three
embeddings of FML into classical higher-order logic (HOL) are
provided side by side: a deep embedding, a heavyweight maximal-shallow
embedding, and a lightweight minimal-shallow embedding. The
minimal-shallow embedding is presented as an Isabelle/HOL locale,
parametrised by an accessibility relation, a world-indexed
interpretation, a universe of worlds, and a variable assignment;
the locale form admits a global faithfulness theorem, stating that
quantifying over all minimal-shallow interpretations recovers
exactly deep validity.

A central technical contribution is a mechanisation, for FML under
constant-domain Kripke semantics, of the (countable) downward
Löwenheim–Skolem theorem, which underpins the automation of our
faithfulness proof between the deep and minimal-shallow embeddings.
Deploying it inside an extension of the minimal-shallow locale
resolves the surjectivity problem that arises against an uncountable
domain of individuals — where the locale's variable assignment,
having countable domain $V = \mathit{nat}$, cannot be surjective
onto the domain — and thereby yields faithfulness over the full
domain.  The construction saturates the individual domain and the
world universe simultaneously, so it applies to arbitrary, possibly
uncountable, domains of both individuals and worlds.

Since prior work treats only propositional logics, with
quantification at most over truth values, we develop here the substitution machinery (free/bound-variable predicates, the
fresh-variable function, capture-avoiding substitution, alphabetic
renaming, the substitutability predicate, the substitution lemma,
and size-based induction principles) needed for the first-order
quantifiers.
\end{abstract}


\maketitle

\section{Motivation and Introduction}

First-order modal logic (FML) sits at the intersection of two of the most pervasive sources of expressive power in symbolic logic: quantification and modality. It is a natural setting for many applications of logic in artificial intelligence, computer science, and philosophy. Embeddings into classical higher-order logic (HOL)~\cite{Andrews02,church1940} have repeatedly been shown to provide a fruitful host environment for such logics; see~\cite{J48,J75} and the references therein. 

\emph{Shallow} embeddings, which identify object-level formulae with HOL terms via the standard translation, give immediate access to Isabelle/HOL's automation, both proof search via \texttt{sledgehammer} and countermodel finding via \texttt{nitpick}. \emph{Deep} embeddings, in which formulae are represented as elements of an inductively defined datatype, are needed for metalogical reasoning about syntax, substitution, or proof systems.

We have recently shown~\cite{C98} that \emph{both} approaches can be carried out simultaneously, in a single HOL theory, in such a way that mutual faithfulness proofs can be automated. That paper focuses on propositional modal logic (PML), where modality is the only complication; its journal version~\cite{JAIR-DeepShallow} adds quantified Boolean formulas, whose quantifiers range over the two truth values only. The natural next step, which we take in this paper, is to add first-order quantification over arbitrary domains in order to move from PML to FML.

\paragraph{Contributions.}
We contribute \textsf{MinFMLinHOL},\footnote{The \textsf{Min} prefix stands for \emph{minimal}; the reasons will become clear below.} an Isabelle/HOL development covering FML with binary relational atoms, propositional connectives, existential quantification, and a normal modality, under constant-domain Kripke semantics. \begin{enumerate}
\item We give a deep embedding (datatype $\mathcal{F}$) and two
shallow embeddings, the maximal one carrying explicit dependencies
on universe~$W$, accessibility~$R$, interpretation~$I$,
domain~$D$ of individuals, variable assignment~$g$, and current world~$w$,
and the minimal one carrying only the world dependency.
\item The minimal-shallow embedding is presented as an
Isabelle/HOL \emph{locale} \texttt{MinS}, parametrised by $WW$, $RR$, $II$, $gg$.
Compared with the constants-only presentation
of~\cite{C98}, the locale form admits a strong
global faithfulness theorem \textsf{FaithfulMS\_all} which states
that quantifying over all locale interpretations recovers exactly
deep validity.  Locale-internal proofs transfer to deep validity by
a single application of this theorem, demonstrated in
\S\ref{sec:exp-locale}.
\item We develop the substitution machinery (free/bound/fresh-variable
predicates, capture-avoiding substitution, alphabetic renaming,
substitutability predicate, the substitution lemma, and size-based
induction principles) for the deep embedding of FML.  This is new
relative to~\cite{C98}, which treats only the propositional fragment
and so contains no substitution machinery, and it goes beyond the
variable-for-variable substitution used for QBF
in~\cite{JAIR-DeepShallow}.  The modal operator is
transparent with respect to variable binding, so every
free/bound/fresh predicate, every substitution and renaming function,
and every induction principle extends by a single uniform modal clause.
\item We mechanise, for FML under constant-domain Kripke semantics,
the (countable) \emph{downward L\"owenheim--Skolem theorem} (theorem
\textsf{DownwardLowenheimSkolem}): every constant-domain Kripke model
has a countable elementary sub-model --- a countable sub-domain of
individuals together with a countable set of worlds --- through which
truth in the deep embedding is preserved.  This resolves the \emph{surjectivity
problem} in relating the deep and minimal-shallow embeddings: the
locale's variable assignment $gg : V \to D$ has countable range and so
cannot be surjective onto an uncountable~$D$; consequently the
deep$\leftrightarrow$minimal-shallow faithfulness lemma reaches only
the assignment's countable image \texttt{Range~gg}, not the full
domain~$D$.  L\"owenheim--Skolem lets
us replace~$D$ by a countable elementary sub-domain onto which $gg$
\emph{can} be made surjective, lifting faithfulness to the full
domain --- yielding the strong global results \textsf{FaithfulMS\_all}
and its full-domain variant \textsf{FaithfulMS\_all'}.
\item We illustrate the development with experiments confirming the
K-axiom, the necessitation rule, the Barcan and converse Barcan
formulae (both valid here under constant domain), and frame
correspondences between reflexive/transitive/symmetric frames and
the T, 4, and B axiom schemata.
\end{enumerate}

\paragraph{Organisation.}
\S\ref{sec:hol} fixes notation for HOL. \S\ref{sec:fml} recalls FML in standard textbook form. \S\S\ref{sec:deep}--\ref{sec:minshallow} present the three embeddings; the minimal embedding is given as a locale (\S\ref{sec:minshallow}), following the design principles laid down for PML in~\cite{C98}. \S\ref{sec:subst} outlines the substitution machinery, \S\ref{sec:elem-subst} mechanises the countable downward L\"owenheim--Skolem theorem, and \S\ref{sec:faithful} establishes the faithfulness theorems. \S\ref{sec:exp} reports on experiments. \S\ref{sec:related} discusses related work, and \S\ref{sec:concl} concludes.

\section{Classical higher-order logic}\label{sec:hol}

We adopt the presentation of HOL given
in~\cite{C98} without modification: simply typed
$\lambda$-calculus over a base type~$o$ of truth values and a base
type~$i$ of individuals, Henkin semantics, and the standard
denotation $\llbracket\cdot\rrbracket$ derived from a frame
$\mathcal{D}$ and an interpretation~$I$.  The validity relation
$\models_{\mathrm{HOL}}$ is sound and complete for Henkin models, and
Isabelle/HOL~\cite{Isabelle} provides an implementation of HOL that
suffices for the entire development that follows.

\section{First-order modal logic (FML)}\label{sec:fml}
The syntax of FML, given a (nonempty) signature $R$ of binary
relation symbols and a denumerable set $V$ of individual variables,
is fixed by the grammar
\begin{equation*}\label{eq:syntax}
\varphi,\psi := r(x,y) \mid \neg\varphi \mid \varphi\wedge\psi
                \mid \exists x.\varphi \mid \Diamond\varphi
\end{equation*}
for relation symbols $r\in R$ and individual variables $x,y\in V$.
The further connectives $\vee$, $\supset$, $\forall$, $\Box$
are defined as usual.

A \emph{constant-domain Kripke model} is a tuple
$\langle W,R,I,D\rangle$, where $W$ is a set of possible worlds,\footnote{Textbook presentations usually require $W$ nonempty; our mechanisation need not. The universe~$W$ is a \emph{predicate} over a nonempty world type~$w$, and validity closes over \emph{all} such predicates: the empty one contributes only the vacuously true $\forall w{\in}\emptyset.(\cdots)$, so it neither trivialises validity nor need be excluded. Accordingly, we verify as a theorem (\texttt{ValD\_nonemptyW}) that the variant of validity carrying an explicit nonemptiness premise on~$W$ coincides with our definition; for~$D$ the guard $g\,\mathtt{into}\,D$ already voids the empty case.}
$R$ is a binary accessibility relation on~$W$, $D$ is a fixed
nonempty domain of individuals, and the interpretation $I$ assigns
to each relation symbol $r$ from the signature and each $w\in W$ a
binary relation $I(r)(w)\subseteq D\times D$ on the domain.  A
\emph{variable assignment} $g\colon V\to D$ assigns a domain
element to each individual variable; $g[x\leftarrow d]$ is the
variable assignment that agrees with $g$ except that it sends
$x$ to~$d$.

Truth in a model is given recursively at each world by
\begin{align*}
\langle W,R,I,D\rangle,g,w &\models r(x,y)
        && \text{iff } I(r)(w)(g(x),g(y))\\
\langle W,R,I,D\rangle,g,w &\models \neg\varphi
        && \text{iff not } \langle W,R,I,D\rangle,g,w \models \varphi\\
\langle W,R,I,D\rangle,g,w &\models \varphi\wedge\psi
        && \text{iff } \langle W,R,I,D\rangle,g,w \models \varphi
            \text{ and } \langle W,R,I,D\rangle,g,w \models \psi\\
\langle W,R,I,D\rangle,g,w &\models \exists x.\varphi
        && \text{iff } \exists d\in D.\, \langle W,R,I,D\rangle,g[x\leftarrow d],w \models \varphi\\
\langle W,R,I,D\rangle,g,w &\models \Diamond\varphi
        && \text{iff } \exists v\in W.\, R(w,v) \wedge \langle W,R,I,D\rangle,g,v \models \varphi
\end{align*}
A formula~$\varphi$ is \emph{valid} iff
$\langle W,R,I,D\rangle,g,w\models\varphi$ for every model
$\langle W,R,I,D\rangle$, every assignment~$g$ into $D$, and every world
$w\in W$.  As usual, frame conditions on the accessibility relation
correspond to modal axioms: reflexivity to T, transitivity to 4,
symmetry to B, and so on (see \S\ref{sec:exp} for our mechanised
versions).  We adopt constant-domain rigid-designator semantics. \emph{Constant domain}: one domain~$D$ is shared by all worlds,
rather than a world-indexed family $(D_w)_{w\in W}$, so the same
individuals exist everywhere and both Barcan formulas are valid
(\S\ref{sec:exp}). \emph{Rigid designation}: a variable denotes the
same individual at every world, i.e.\ the assignment $g\colon V\to D$
is world-independent (not $g\colon V\to W\to D$); this is why a single
$g$ threads through the truth clauses, untouched by the modal shift.
Relaxing either is the varying-domain extension left to future work
(\S\ref{sec:related}).

\paragraph{Relation to the basic first-order modal logic of the literature.}
The setting just described is the constant-domain basic first-order
modal logic studied by Br\"auner and
Ghilardi~\cite[\S2]{BraunerGhilardi2007}.  For reasons of minimality
we deviate from their formulation in four respects, none of which
affects the modal-logical content of the system.  First, our signature contains only \emph{binary} relation symbols,
whereas the cited reference admits predicate symbols of arbitrary
arity (including 0-place symbols, i.e.\ propositional letters).
Predicate symbols of other arities can be encoded in our binary
fragment: unary atoms $P(x)$ via $P'(x,x)$ or $P'(x,d_0)$ for a
fixed parameter, and $n$-ary atoms ($n\geq 3$) by standard
tuple-reification (reifying each $n$-tuple as an individual with projection relations).  We do not pursue these encodings
here, since the binary fragment already exhibits all the
modal-logical phenomena  we are interested in. Second, we omit the primitive equality predicate $x = y$
that the cited reference includes; equality strictly increases
expressive power and is not definable from a generic binary
relation symbol alone.  Extending our setting with equality is
routine---one adds an identity clause to the interpretation~$I$ and
the obvious recursion clause
$\langle W,R,I,D\rangle,g,w\models x = y$ iff $g(x)=g(y)$---but
nothing in the present development depends on it.  Third, we take
$\exists$ and $\Diamond$ as primitives and define $\forall, \Box$
by duality, whereas \cite{BraunerGhilardi2007} proceeds with
$\forall, \Box$; this is a purely presentational choice with no
semantic content.  Taking $\Diamond$ and $\exists$ as primitives keeps the
L\"owenheim--Skolem construction of \S\ref{sec:elem-subst} simple:
it is carried out directly in \emph{existential} form, where the
quantifier and the modality are handled uniformly---each merely
calls for a witness (an individual for $\exists$, an accessible
world for $\Diamond$).  The dual pair $\forall,\Box$ would be no less
uniform, but the witness construction is markedly more direct in the
existential formulation.  Fourth, and likewise, we take $\neg,\wedge$ as primitive and derive
$\vee,\supset$, rather than starting from $\supset$ and a falsum~$\bot$.
As the host logic is \emph{classical} HOL the Boolean base is
immaterial: $\neg,\wedge$ and $\supset,\bot$ are interderivable
pointwise under the deep truth predicate, a routine check by unfolding
the definitions.  The base would matter for an \emph{intuitionistic} object
logic, where $\{\supset,\bot\}$ is primitive; such a logic still
embeds shallowly in HOL via its (monotone) Kripke
semantics~\cite{J21}, so an intuitionistic-modal variant is feasible
future work rather than a mere change of connectives
(cf.~\cite{KirstShillito2025}, \S\ref{sec:related}).

Subject to these scope restrictions, the semantics adopted
here---constant domains, rigid variables, a single Kripke
accessibility relation, and the standard clauses for $\exists$ and
$\Diamond$---coincides with the constant-domain basic semantics
of~\cite[\S2.2]{BraunerGhilardi2007}, so the familiar
frame-correspondence facts for the axioms T, 4, B and for the
Barcan / converse Barcan formulas apply unchanged.
\section{Deep embedding of FML in HOL}\label{sec:deep}

\begin{figure}[htbp]
  \centering
  \colorbox{gray!30}{\includegraphics[width=.95\columnwidth]{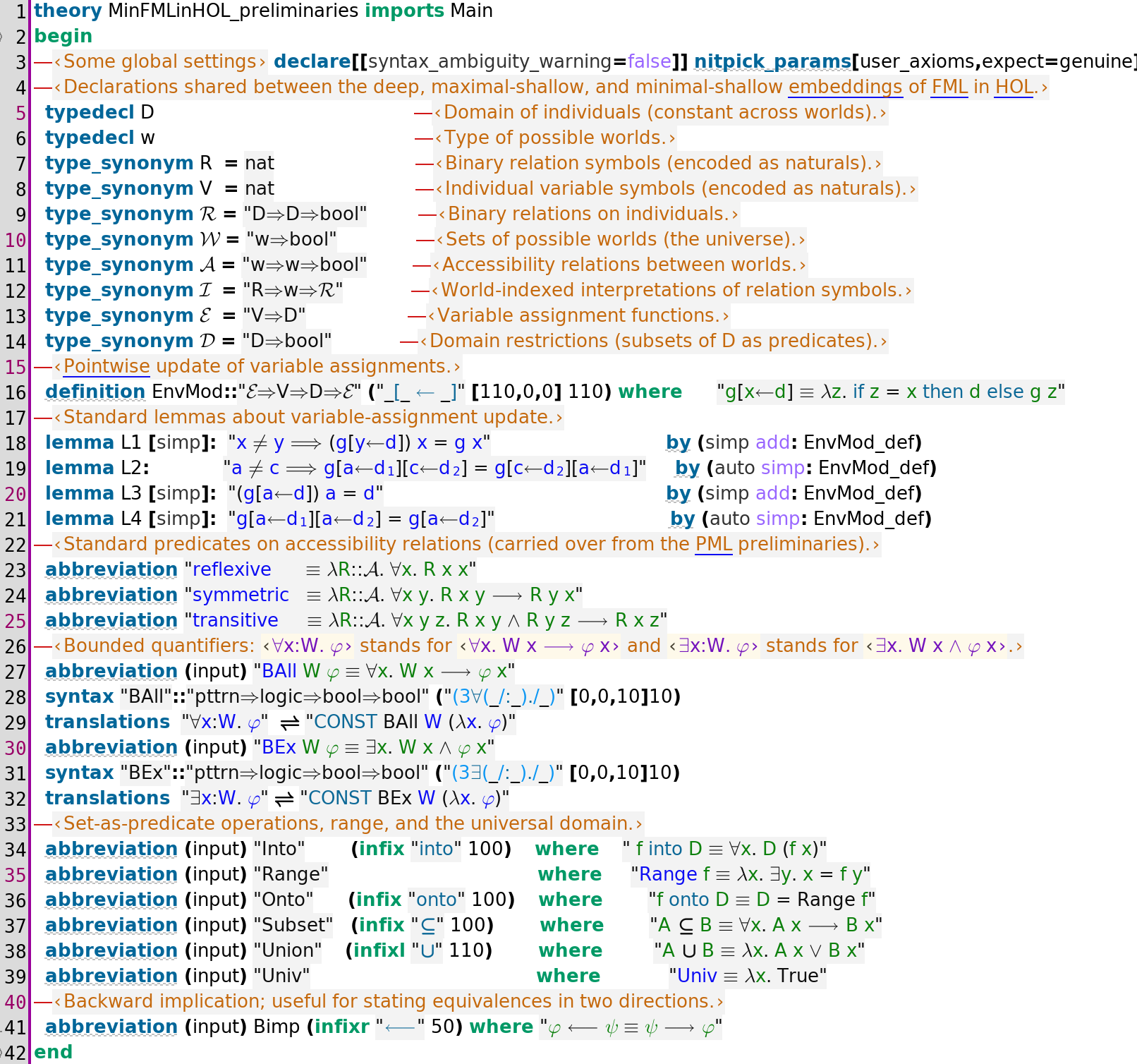}}
  \caption{The shared preliminaries theory
  \texttt{MinFMLinHOL\_preliminaries.thy}: type declarations
  (lines~5--14), the variable-assignment update operator
  $g[x{\leftarrow}d]$ (line~16) with companion lemmata
  \texttt{L1}--\texttt{L4} (lines~18--21), the accessibility-relation
  predicates \texttt{reflexive}, \texttt{symmetric},
  \texttt{transitive} (lines~23--25), the bounded quantifiers
  $\forall x{:}W$, $\exists x{:}W$ (lines~27--32), and set-as-predicate
  operations \texttt{into}, \texttt{Range}, \texttt{onto},
  $\sqsubseteq$, $\sqcup$, \texttt{Univ} (lines~34--39).
  \label{fig:MinFMLinHOL_preliminaries}}
\end{figure}

Preliminaries (Figure~\ref{fig:MinFMLinHOL_preliminaries}) shared by all embeddings declare the types $D$
(individuals) and $w$ (worlds), together with the synonyms
$R:=\mathit{nat}$ (relation symbols), $V:=\mathit{nat}$ (variables, via
natural numbers, with substitution and renaming developed in
\S\ref{sec:subst}),
$\mathcal{R}:=D\Rightarrow D\Rightarrow\mathit{bool}$,
$\mathcal{W}:=w\Rightarrow\mathit{bool}$,
$\mathcal{A}:=w\Rightarrow w\Rightarrow\mathit{bool}$,
$\mathcal{I}:=R\Rightarrow w\Rightarrow\mathcal{R}$,
$\mathcal{E}:=V\Rightarrow D$, and a domain restriction
$\mathcal{D}:=D\Rightarrow\mathit{bool}$.  Variable-assignment update
$g[x\leftarrow d]$ (line~16 of the figure), the bounded quantifiers
$\forall x{:}W$, $\exists x{:}W$ (lines~27--32), and set-as-predicate operations
$\sqsubseteq$, $\sqcup$, \texttt{Range}, \texttt{Univ},
\texttt{into}, \texttt{onto} (lines~34--39) are introduced in the
theory file \texttt{MinFMLinHOL\_preliminaries.thy} and used by all
subsequent files (in particular by the L\"owenheim--Skolem proof of
\S\ref{sec:elem-subst}).

\begin{figure}[htbp]
  \centering
  \colorbox{gray!30}{\includegraphics[width=.95\columnwidth]{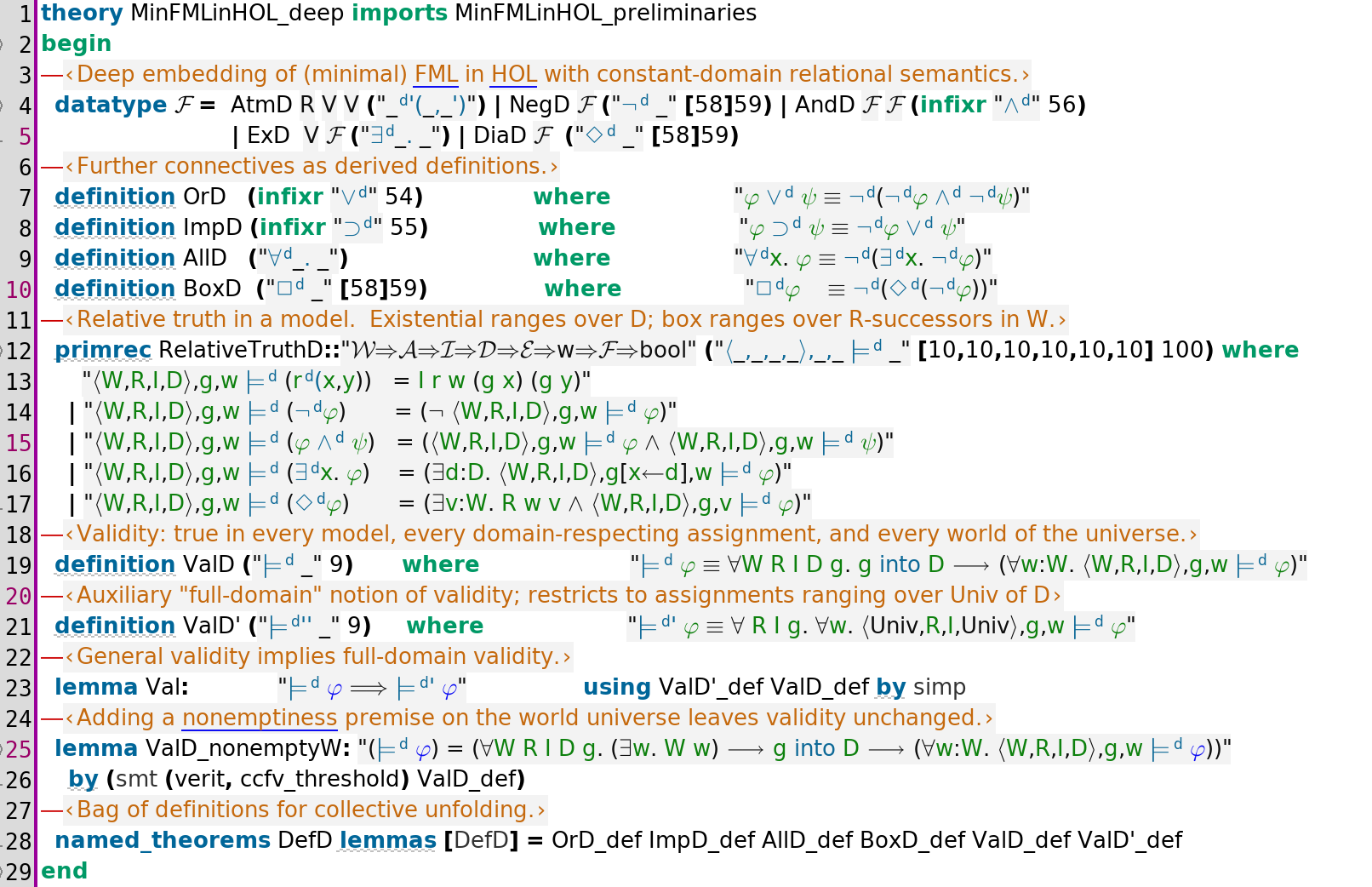}}
  \caption{Deep embedding of FML in HOL: the datatype $\mathcal{F}$ of formulae
  with derived connectives, the relative-truth predicate
  \texttt{RelativeTruthD} (lines~12--17), the validity definitions
  \texttt{ValD}, \texttt{ValD'} (lines~19--21), and the
  nonemptiness-invariance lemma \texttt{ValD\_nonemptyW}
  (lines~24--26).
  \label{fig:MinFMLinHOL_deep}}
\end{figure}

The deep embedding is a single inductive datatype $\mathcal{F}$ whose
constructors mirror the grammar~\eqref{eq:syntax} (theory
\texttt{MinFMLinHOL\_deep.thy}, Figure~\ref{fig:MinFMLinHOL_deep},
lines~4--5; derived connectives on lines~7--10).
The relative-truth predicate
$\langle W,R,I,D\rangle,g,w \models^d \varphi$ is declared using mixfix syntax and defined by primitive recursion (lines~12--17): the existential quantifier
ranges over the explicit domain~$D$ (only those $d$ with $D\,d$);
the diamond ranges over $R$-successors inside the universe~$W$ (the
box being derived by duality); the rest is standard.
Validity \texttt{ValD} (line~19) closes universally over models, over
domain-respecting assignments, and over worlds of the universe.

\paragraph{Universe and explicit domain.}
The deep embedding's validity makes two additional parameters explicit compared
to the propositional embedding in~\cite{C98}: the universe $W$ of
worlds (present in some form in any modal embedding) and the
domain $D$ of individuals over which the existential quantifier
ranges (which has no analogue in the propositional setting).  Both
are represented here as HOL predicates, and \texttt{ValD} closes universally over
models, over assignments lying in the domain (\texttt{g into D}),
and over worlds in the universe ($\forall w{:}W$).  The minimal-shallow embedding (\S\ref{sec:minshallow}) keeps $WW$
as an explicit locale parameter to maintain a uniform dependency
across all three embeddings.  This has two practical benefits: the
faithfulness theorems share the same parameters across the
embeddings, and counterexamples found by \texttt{nitpick} display
the universe restriction explicitly, matching the deep embedding's
presentation.

\paragraph{Auxiliary full-domain validity.}
A second validity definition $\models^{d'}$ restricts to variable
assignments ranging over the universal domain \texttt{Univ} (the
predicate $\lambda x.\,\mathit{True}$, see
Fig.~\ref{fig:MinFMLinHOL_preliminaries}, line~39) on the type~$D$
of individuals, and likewise takes \texttt{Univ} as the universe of
worlds.  This is a convenience for \texttt{nitpick}, which
handles the trivial \texttt{Univ} predicate more efficiently than a
guarded $D\,x$ side condition on every binder.  Since
$\models^d\varphi$ implies $\models^{d'}\varphi$, any counterexample
found by \texttt{nitpick} for $\models^{d'}$ refutes $\models^d$ as
well.  Moreover, the apparent restriction to $D = \mathtt{Univ}$
does not weaken \texttt{nitpick}'s countermodel-finding power: \texttt{nitpick} is
free to choose any concrete finite domain for the \emph{type} of
individuals, so a counterexample at some smaller predicate $D$ can
equivalently be found by shrinking the type and using the unguarded
$\mathtt{Univ}$ on it.

\section{Maximal shallow embedding}\label{sec:maxshallow}

\begin{figure}[htbp]
  \centering
  \colorbox{gray!30}{\includegraphics[width=.95\columnwidth]{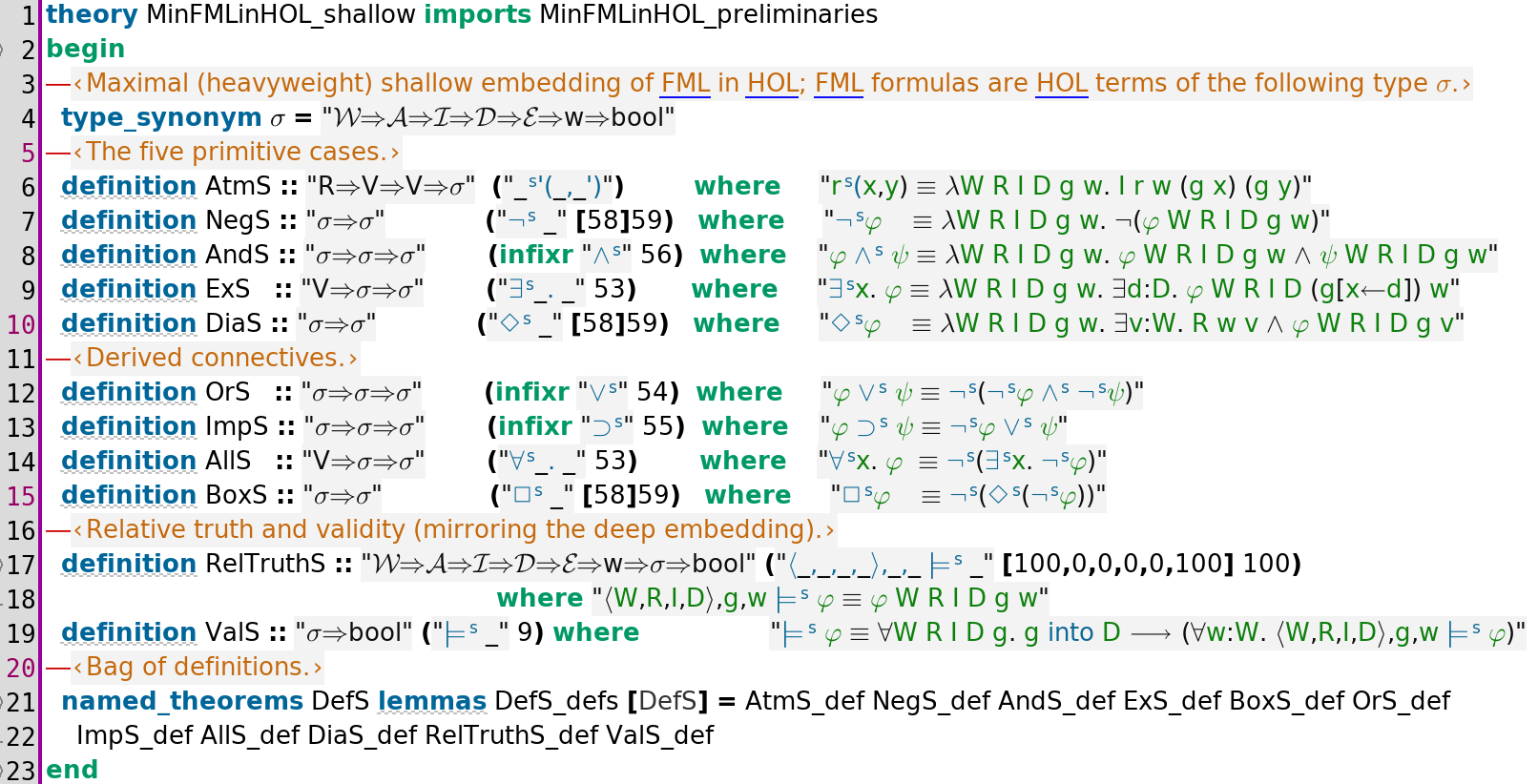}}
  \caption{Maximal shallow embedding of FML in HOL: the formula type
  $\sigma$ (line~4), primitive cases (lines~6--10), derived
  connectives (lines~12--15), and validity (lines~17--19).
  \label{fig:MinFMLinHOL_shallow}}
\end{figure}

The maximal shallow embedding lifts every primitive of FML to a
$\lambda$-term abstracted over the explicit dependencies $W$, $R$,
$I$, $D$, $g$, $w$.  The type of formulae is
$\sigma := \mathcal{W}\Rightarrow\mathcal{A}\Rightarrow
\mathcal{I}\Rightarrow\mathcal{D}\Rightarrow\mathcal{E}\Rightarrow
w\Rightarrow\mathit{bool}$ (theory \texttt{MinFMLinHOL\_shallow.thy},
Figure~\ref{fig:MinFMLinHOL_shallow}, line~4).  Atoms consult the
world-indexed interpretation (line~6), the existential introduces a
meta-level HOL existential paired with a variable-assignment update (line~9),
and the diamond is implemented by a meta-level HOL existential over
accessibility-successors inside~$W$ (line~10).  Derived connectives
are on lines~12--15; relative truth and validity on lines~17--19.
The pattern is identical to that of~\cite{C98} for PML; the only
difference is that the formula type $\sigma$ is more complex,
because both quantifier binding and the modal world-shift must be
accommodated, and because the explicit domain~$D$ has been threaded
through.

\section{Minimal shallow embedding (locale-based)}\label{sec:minshallow}

\begin{figure}[htbp]
  \centering
  \colorbox{gray!30}{\includegraphics[width=.95\columnwidth]{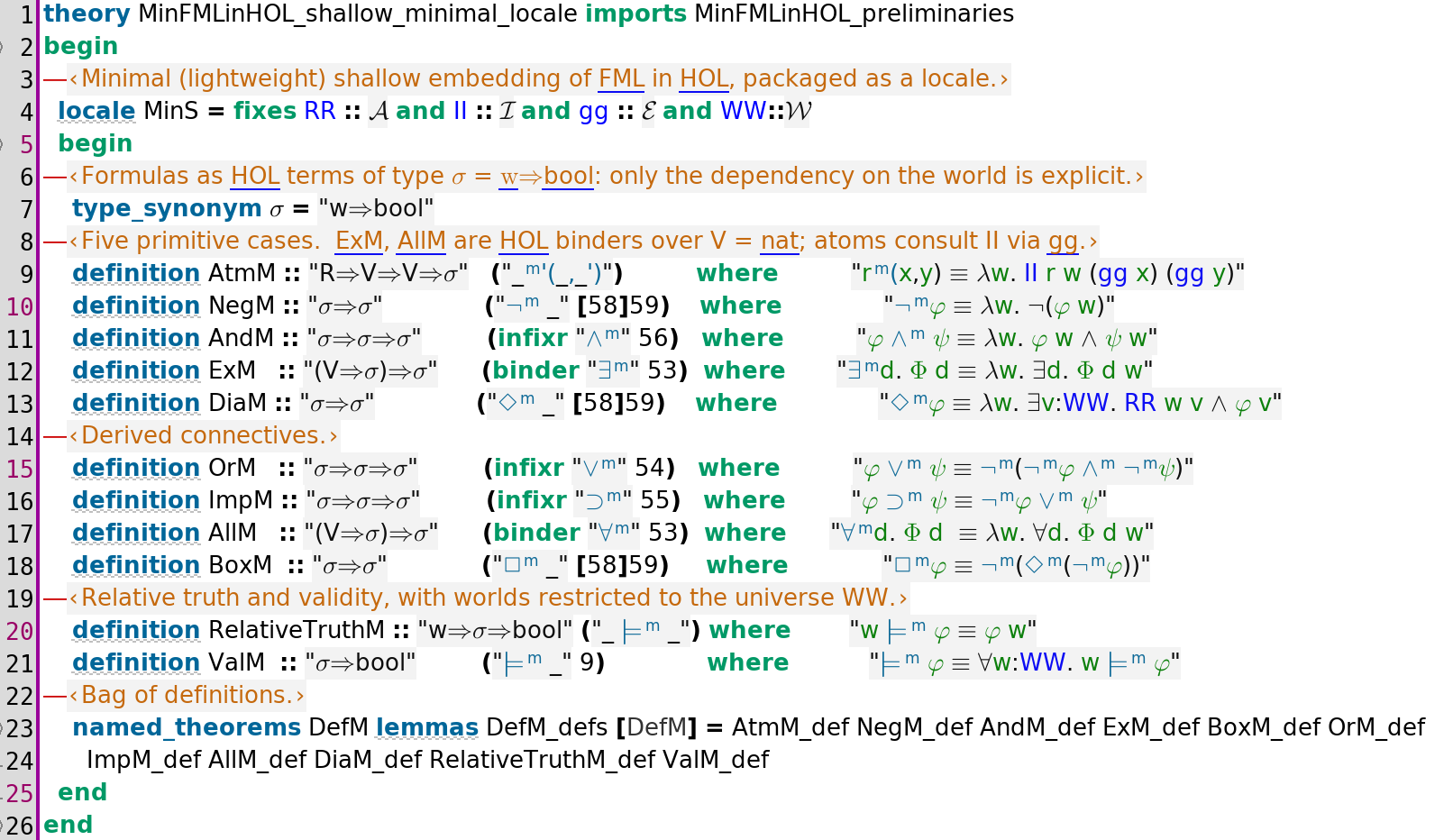}}
  \caption{The locale \texttt{MinS} packaging the minimal shallow
  embedding: parameters $RR$, $II$, $gg$, $WW$ (line~4), formula type
  $\sigma$ (line~7), primitive cases (lines~9--13), and derived
  connectives (lines~15--18).
  \label{fig:MinFMLinHOL_shallow_minimal_locale}}
\end{figure}

The maximal shallow type carries every dependency, but only $g$ (at
the existential) and $w$ (at the modality) ever change during recursion.
Following the design principle of~\cite{C98}, the minimal embedding
turns four of these dependencies---$W$, $R$, $I$, and $g$---into locale
parameters (constants of the meta-logic), leaving the world~$w$ as the
only argument that still varies during recursion. The domain of individuals arises implicitly as the range of the assignment function $g$; the connection to explicit domains~$D$ is deferred to the refinement of
\S\ref{sec:elem-subst}.  Promoting $g$ to a static parameter is
possible because the minimal-shallow existential is encoded directly by
HOL's own binder (\texttt{ExM}, line~12 of
Fig.~\ref{fig:MinFMLinHOL_shallow_minimal_locale}); the
variable-assignment update therefore happens at the meta-level
rather than via a recursive call with an updated $g$, and $gg$ need
only interpret the \emph{free} variables of the input formula.

In contrast to~\cite{C98}, however, we package the minimal embedding
as an Isabelle/HOL \emph{locale}~\cite{ballarin2014}.\footnote{A
\emph{locale} in Isabelle/HOL is a named context that fixes typed
parameters and (optionally) assumes axioms about them; theorems
proved inside the locale automatically apply to every instantiation
of the parameters that satisfies the assumptions.}  The locale form
has two advantages: 
\begin{itemize}
\item different minimal interpretations can coexist within a single
  theory, since the four parameters $(WW, RR, II, gg)$ are first-class
  arguments rather than fixed constants;
\item it admits a strong global theorem, \textsf{FaithfulMS\_all}
  (\S\ref{sec:faithful}), stating that quantifying over all locale
  interpretations recovers exactly deep validity.
\end{itemize}
The locale \texttt{MinS} (theory
\texttt{MinFMLinHOL\_shallow\_minimal\_locale.thy},
Figure~\ref{fig:MinFMLinHOL_shallow_minimal_locale}, line~4) fixes
the four parameters and reduces the formula type to
$w\Rightarrow\mathit{bool}$ (line~7).  The five primitive cases are
on lines~9--13 (\texttt{AtmM}, \texttt{NegM}, \texttt{AndM},
\texttt{ExM}, \texttt{DiaM}), the derived connectives on
lines~15--18, and relative truth and validity on lines~20--21.

Two design choices in this minimal-embedding setup are worth
highlighting.

\paragraph{Binder for \texttt{ExM}.}
\texttt{ExM} (and the derived \texttt{AllM}) is declared as a HOL
binder over the variable type~$V$.  This lets \texttt{sledgehammer}
treat first-order modal goals as ordinary HOL goals: bound variables
under \texttt{ExM} are HOL-level bound variables, so the external
first-order provers behind \texttt{sledgehammer} can reason about
them directly without needing to unfold the deep-embedding syntax.
In the deep embedding, by contrast, \texttt{ExD} carries the
variable name explicitly as part of the syntax tree (a named-variable
binder, names encoded as natural numbers); the bridge between the
two styles is built inside the deep-to-minimal translation
\texttt{DpToShM} of~\S\ref{sec:faithful} (see lines~8--9 of
Fig.~\ref{fig:MinFMLinHOL_faithfulness_locale}) via the
safe-substitution operation $[v\!\leftarrow_r\!d](\varphi)$.

\begin{figure}[htbp]
  \centering
  \colorbox{gray!30}{\includegraphics[width=.95\columnwidth]{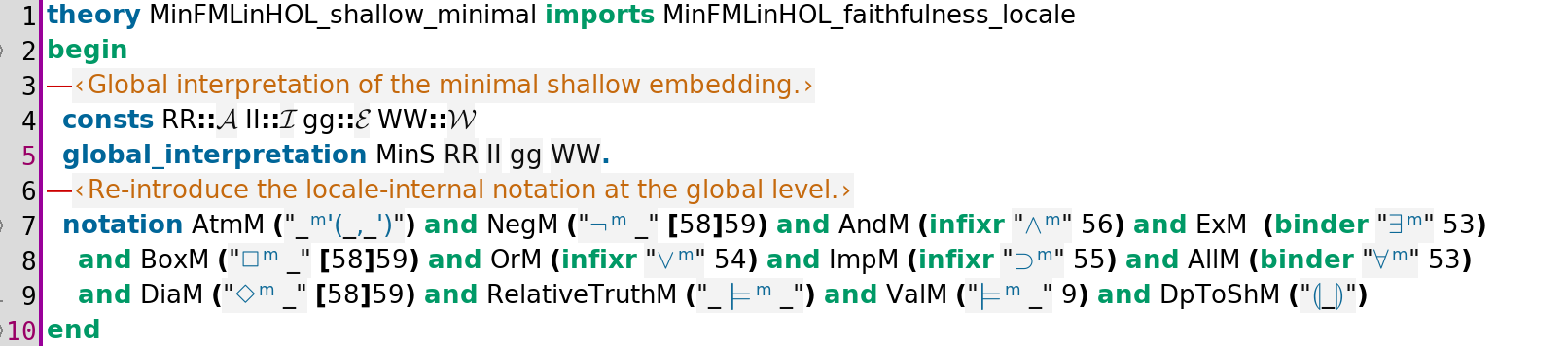}}
  \caption{Global instantiation of the minimal-shallow locale
  (theory \texttt{MinFMLinHOL\_shallow\_minimal.thy}): four
  uninterpreted constants \texttt{RR}, \texttt{II}, \texttt{gg},
  \texttt{WW} (line~4); the
  \texttt{global\_interpretation MinS} statement (line~5);
  re-introduction of the locale's notation at the global level
  (lines~7--9).
  \label{fig:MinFMLinHOL_shallow_minimal}}
\end{figure}

\paragraph{Constants vs.\ locale parameters.}
For practical experimentation, the locale needs to be instantiated
once at the global level for \texttt{nitpick} to construct usable
countermodels: \texttt{nitpick} handles uninterpreted constants more
reliably than locale parameters.  The sibling theory
\texttt{MinFMLinHOL\_shallow\_minimal.thy}
(Figure~\ref{fig:MinFMLinHOL_shallow_minimal}) introduces
uninterpreted constants \texttt{RR}, \texttt{II}, \texttt{WW},
\texttt{gg} (line~4) and performs a \texttt{global\_interpretation}
with them (line~5); this is the standard ``constants-only''
presentation of the minimal embedding, as in~\cite{C98}.  The locale
formulation in \texttt{MinFMLinHOL\_shallow\_minimal\_locale.thy}
remains the single source of all theorems; the global interpretation
propagates them to the constants automatically, without proof
duplication.

\section{Substitution machinery}\label{sec:subst}

The substitution lemma is necessary for the
deep$\leftrightarrow$minimal-shallow faithfulness proof to handle
the existential case.  Since~\cite{C98} treats only propositional
modal logic, no substitution machinery is provided there; we develop
one here, in the theory \texttt{MinFMLinHOL\_deep\_subst\_lemma.thy}.
The components are standard for first-order
languages: a free-variable predicate \texttt{is\_free}, a
bound-variable predicate \texttt{is\_bound}, a
fresh-variable function \texttt{fresh} with
consequence lemmata \texttt{L5}--\texttt{L9}, the
irrelevance lemma \texttt{L12}, variable-for-variable
substitution \texttt{Subst} written $[x{\leftarrow}z](\varphi)$, the
substitutability predicate \texttt{is\_subst\_for}, the
\textsf{SubstitutionLemma} itself, the
alphabetic-renaming function \texttt{ren\_for\_subst},
and the safe-substitution operation \texttt{ren\_subst}, written
$[x\!\leftarrow_r\!z](\varphi)$.  Two size-based induction
principles \texttt{SInduct} and \texttt{QInduct}
 drive
the trickier inductions.

Compared to plain first-order logic,
every recursive definition and every inductive lemma in the chain
is uniformly extended with a clause for the modality.
Because the modal operator binds no variable but merely shifts the
world, the modal case is in every instance a single line of the same
structurally trivial form: it descends transparently, just like the
negation case,
and is dispatched in every lemma between \texttt{L5} and
\texttt{L28} by the same one-line automation (\texttt{auto},
\texttt{metis}, \texttt{smt}).
The two key outcomes are the substitution lemma
\textsf{SubstitutionLemma} and
its safe-substitution form \texttt{L27}; the
latter is what the existential case of the faithfulness proof
eventually invokes.

\paragraph{Coincidence-of-variable-assignments lemma.}
A simple but useful consequence is that relative truth depends only
on the values of the variable assignment on free variables (lemma
\texttt{L1} of \texttt{MinFMLinHOL\_ElementarySubstructure.thy}),
proved by induction.

\section{The countable downward L\"owenheim--Skolem theorem}
\label{sec:elem-subst}

The main result of this section is a mechanisation of the (countable) \emph{downward
L\"owenheim--Skolem theorem} for FML under
constant-domain Kripke semantics (theory
\texttt{MinFMLinHOL\_ElementarySubstructure.thy}).
It is used to resolve the \emph{surjectivity problem}
that arises when relating the deep and minimal-shallow embeddings
over an arbitrary --- possibly uncountable --- domain of
individuals~$D$ and universe of worlds~$W$.

\paragraph{The theorem.}
For every constant-domain Kripke model $\langle W,R,I,D\rangle$, every
countable initial sub-domain $D_0\sqsubseteq D$, and every countable
initial set of worlds $W_0\sqsubseteq W$, there exists a countable
elementary sub-model $\langle\mathcal{W},R,I,\mathcal{N}\rangle$ with
$D_0\sqsubseteq\mathcal{N}\sqsubseteq D$ and
$W_0\sqsubseteq\mathcal{W}\sqsubseteq W$, both $\mathcal{N}$ and
$\mathcal{W}$ countable, through which truth in the deep embedding is
preserved.\footnote{Here $\mathcal{N}$ and $\mathcal{W}$ abbreviate
$\mathcal{N}(D_0,W_0,W,R,I,D)$ and $\mathcal{W}(D_0,W_0,W,R,I,D)$.}
Theorem \textsf{DownwardLowenheimSkolem} states that
there exist sub-predicates $N$ and $W'$ with
$D_0\sqsubseteq N\sqsubseteq D$ and $W_0\sqsubseteq W'\sqsubseteq W$,
both countable, and with
$\langle W,R,I,D\rangle,g,w\models^d\varphi
=\langle W',R,I,N\rangle,g,w\models^d\varphi$ for every $g$ ranging
into~$N$, every world~$w$ with $W'\,w$, and every formula~$\varphi$.
Its proof instantiates $N:=\mathcal{N}$ and $W':=\mathcal{W}$ and
combines the containment and countability lemmata for $\mathcal{N}$
and $\mathcal{W}$ with the elementary core
$\mathcal{N}\_valid$, the sub-model
$\langle\mathcal{W},R,I,\mathcal{N}\rangle$ being constructed below.

\paragraph{Why this resolves the surjectivity problem.}
In the minimal-shallow embedding the variable assignment
$gg : V \to D$ has $V = \mathit{nat}$, so its image is countable.
The deep embedding, however, lets the existential range over the
entire domain~$D$.  When~$D$ is uncountable, no variable assignment
of this form can be surjective onto~$D$; consequently faithfulness
over the full domain~$D$ does not follow immediately from
\textsf{FaithfulMDlem}, whose statement is restricted to
\texttt{Range~gg}.\footnote{The restriction is due to the use of substitution for the translation of the existential, which is necessary since the assignment $gg$ is a fixed parameter.}  The L\"owenheim--Skolem theorem lets us
replace~$D$ by a countable elementary sub-domain~$\mathcal{N}$ (and
$W$ by a countable~$\mathcal{W}$) onto which $gg$ \emph{can} be made
surjective; faithfulness against~$D$ is then recovered by transferring
truth between the full model and the elementary sub-model via
\textsf{DownwardLowenheimSkolem}.  This transfer is packaged by the
locale \texttt{MinS\_ES} and applied in
\S\ref{sec:faithful}.

\paragraph{Construction.}
By the \emph{Tarski--Vaught test}, a sub-model
$\langle W',R,I,D'\rangle$ with $D'\sqsubseteq D$ and
$W'\sqsubseteq W$ is elementary iff (a)~every existential
$\exists^d y.\,\varphi$ satisfied by an assignment into~$D'$ at a world
of~$W'$ has an individual witness in~$D'$, and (b)~every diamond
$\Diamond^d\varphi$ satisfied at a world of~$W'$ has an $R$-successor
witness in~$W'$ (lemma \texttt{TarskiVaught}).  It
therefore suffices to construct countable $\mathcal{N}\sqsubseteq D$
and $\mathcal{W}\sqsubseteq W$ closed under both kinds of witness.
Because the deep primitive is the diamond (\S\ref{sec:deep}), the box
is handled by duality and needs no separate treatment.  The
construction proceeds in three steps.

\subparagraph{(i) Witnesses.}
Two Hilbert-choice operators pick witnesses.  For a partial
assignment~$g$, world~$w$, variable~$y$ and formula~$\varphi$,
\texttt{existential\_witness} returns an individual
in~$D$ witnessing $\exists^d y.\,\varphi$ when that formula holds
at~$w$ under~$g$; dually, \texttt{modal\_witness}
returns an $R$-successor world in~$W$ witnessing $\Diamond^d\varphi$.
Collecting these over all formulae and all partial assignments ranging
into a given sub-domain~$D_0$ at worlds of a given~$W_0$ yields the
witness sets \texttt{existential\_witnesses} and
\texttt{modal\_witnesses}.  The auxiliary
\texttt{restricted\_assigns} encodes the relevant
partial maps --- from the finitely many free variables of~$\varphi$
into~$D_0$ --- as \texttt{map\_of} applied to finite lists, so that
both witness sets inherit countability from~$D_0$ and~$W_0$
(\texttt{existential\_witnesses\_countable},
\texttt{modal\_witnesses\_countable}).

\subparagraph{(ii) Simultaneous stage-wise saturation.}
The elementary sub-model is the limit of an iterated witness
collection seeded by~$D_0$ and~$W_0$.  The stages $\mathcal{N}_i$
(individuals) and $\mathcal{W}_i$ (worlds) are defined by \emph{mutual}
primitive recursion on the stage index: each
$\mathcal{N}_{i+1}$ adds the existential witnesses, and each
$\mathcal{W}_{i+1}$ the modal witnesses, generated at assignments and
worlds ranging into the previous stages $\mathcal{N}_i,\mathcal{W}_i$.
The limits $\mathcal{N}$ and $\mathcal{W}$ are the predicate-level
unions over all stages.  By induction on the stage
index, every stage is countable whenever $D_0$ and $W_0$ are, so
$\mathcal{N}$ and $\mathcal{W}$ are countable as
countable unions of countable sets (\texttt{countable\_N},
\texttt{countable\_W}).  The subset lemmata show
$\mathcal{N}\sqsubseteq D_0\sqcup D$ and
$\mathcal{W}\sqsubseteq W_0\sqcup W$ unconditionally; under
$D_0\sqsubseteq D$ and $W_0\sqsubseteq W$ these
collapse to $\mathcal{N}\sqsubseteq D$ and $\mathcal{W}\sqsubseteq W$,
so the construction stays inside the original model.

\subparagraph{(iii) Tarski--Vaught test.}
The theorem $\mathcal{N}\_valid$ states that truth in
the sub-model $\langle\mathcal{W},R,I,\mathcal{N}\rangle$ agrees with
truth in the full model, i.e.\ that
$\langle\mathcal{W},R,I,\mathcal{N}\rangle$ is an elementary sub-model
of $\langle W,R,I,D\rangle$; this elementary-substructure relation is
written $\langle W',R',I',D'\rangle\subseteq_E\langle W,R,I,D\rangle$
(\texttt{ElementarySubstructure}).  Its proof applies the
Tarski--Vaught test~\texttt{TarskiVaught} and discharges its two
witness conditions by construction: the
existential (resp.\ modal) witness produced by Hilbert's choice is
collected into the next domain (resp.\ world) stage, and hence into
$\mathcal{N}$ (resp.\ $\mathcal{W}$).  Witnesses are constructed
relative to a finite restriction of the assignment to the free
variables of~$\varphi$ --- the
\texttt{restrict\_assign}/\texttt{unrestrict} pair ---
and the coincidence-of-assignments lemma~\texttt{L1}
lifts them to the full assignment.

\paragraph{Locale-level packaging.}
Two locales package the result for the faithfulness proof.  The locale
\texttt{MinS\_ES} extends \texttt{MinS} with a world
predicate~$WW'$ and a domain predicate~$DD$ together with the
assumption
$\langle WW,RR,II,\texttt{Range }gg\rangle\subseteq_E\langle WW',RR,II,DD\rangle$;
its lemma $\mathcal{N}\_valid\_ES$ is the operational
reading of the L\"owenheim--Skolem theorem inside the locale, reading
truth at the full domain~$DD$ off truth at \texttt{Range~gg}.  The
specialisation \texttt{MinS\_ES\_Univ} fixes
$WW'=DD=\texttt{Univ}$, i.e.\ the universal world and individual
domains.  Lemma \texttt{gg\_ex} establishes that, for
any nonempty countable seed, there exists a variable assignment
surjective onto~$\mathcal{N}$; the elementary sub-model itself is the
one furnished by \textsf{DownwardLowenheimSkolem} above.
Finally, the locale \texttt{MinS\_specific} packages
the particular case powering global faithfulness: for a given~$g_0$,
world~$w$ and target formula~$\varphi$ it constructs a bounded
accessibility relation~$RR$ and a surjective $gg$ agreeing
with $g_0$ on the free variables of~$\varphi$ (the passage from the
bounded~$RR$ back to~$R_0$ uses that deep truth depends on the
accessibility relation only up to the modal depth of~$\varphi$, lemma
\texttt{L4} with the $n$-step relation-equivalence predicate),
culminating in the lemma \texttt{valid\_specific},
which transfers
$\langle W_0,R_0,I_0,D_0\rangle,g_0,w\models^d\varphi$ into the
constructed elementary-substructure setting.  This in turn powers the
global theorems \textsf{FaithfulMS\_all} and \textsf{FaithfulMS\_all'}
of \S\ref{sec:faithful}.

\section{Faithfulness, automated}\label{sec:faithful}

\begin{figure}[htbp]
  \centering
  \colorbox{gray!30}{\includegraphics[width=.95\columnwidth]{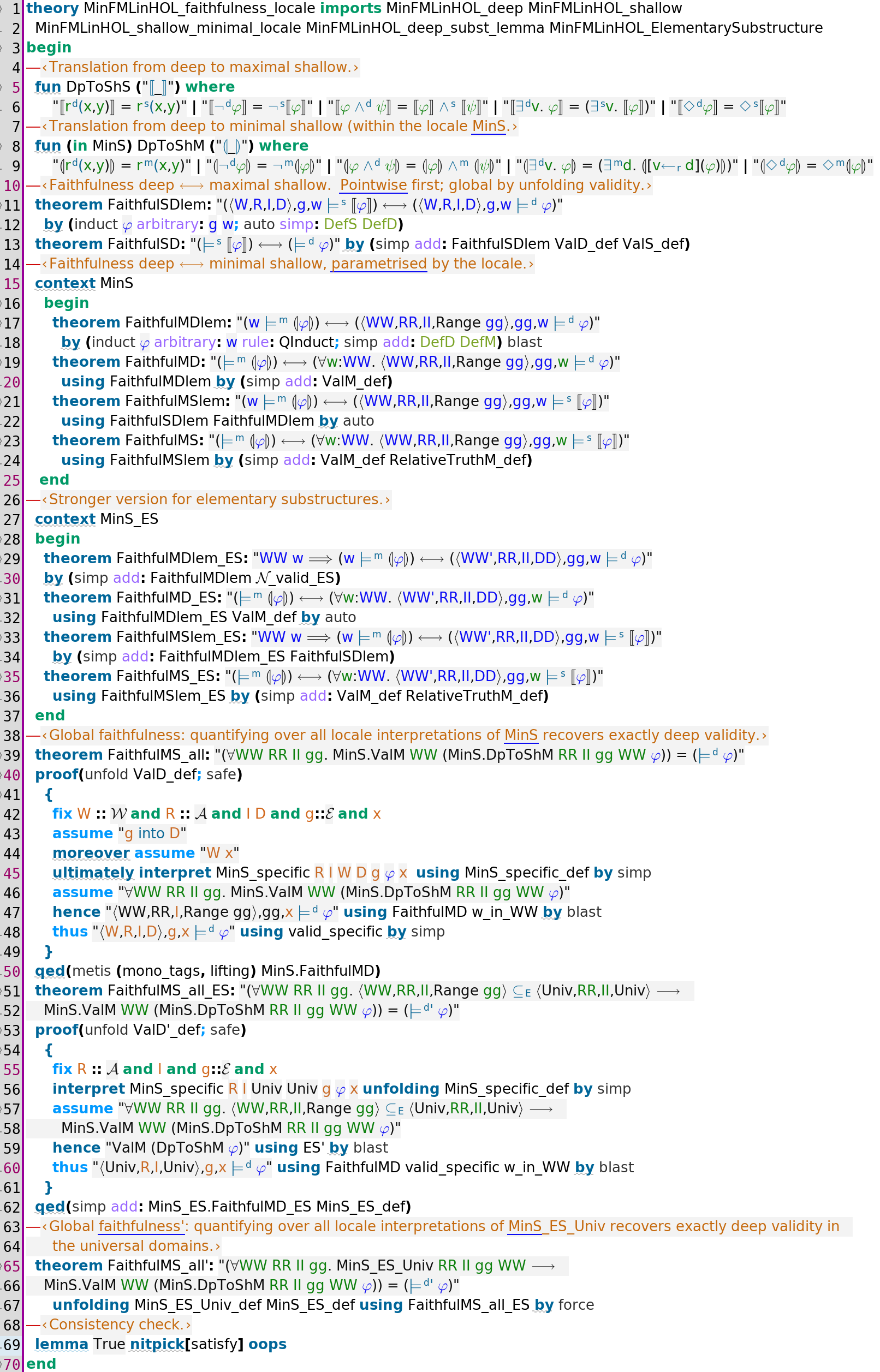}}
  \caption{Faithfulness theorems (theory
  \texttt{MinFMLinHOL\_faithfulness\_locale.thy}): the translations
  $\llbracket\cdot\rrbracket$ and $(\!\lvert\cdot\rvert\!)$
  (lines~5--9), the deep$\leftrightarrow$maximal-shallow
  faithfulness theorems (lines~11--13), the locale-based
  deep$\leftrightarrow$minimal-shallow correspondences
  (\textsf{FaithfulMDlem}, \textsf{FaithfulMD}, \textsf{FaithfulMSlem},
  \textsf{FaithfulMS}, lines~17--24), the stronger versions inside
  \texttt{MinS\_ES} (lines~27--37), and the
  strong global theorems \textsf{FaithfulMS\_all} and
  \textsf{FaithfulMS\_all'} (lines~39, 65; the intermediate
  \textsf{FaithfulMS\_all\_ES} is on lines~51--62).
  \label{fig:MinFMLinHOL_faithfulness_locale}}
\end{figure}

Two mappings translate deep formulae into shallow ones (theory
\texttt{MinFMLinHOL\_faithfulness\_locale.thy},
Fig.~\ref{fig:MinFMLinHOL_faithfulness_locale}):
\texttt{DpToShS}, written $\llbracket\cdot\rrbracket$, on
lines~5--6, and \texttt{DpToShM}, written
$(\!\lvert\cdot\rvert\!)$, on lines~8--9.  The first recurses
transparently through every primitive of FML, since both source and
target use named-variable existentials.  The second,
\texttt{DpToShM}, is declared inside the locale \texttt{MinS}; its
existential clause (line~9) inserts a safe substitution
$[v\!\leftarrow_{r}\!d](\varphi)$ that replaces the named
variable~$v$ in $\varphi$ with the HOL-bound~$d$, bridging
\texttt{ExD}'s named-variable representation and \texttt{ExM}'s
HOL-binder representation.

\paragraph{Pointwise lemmata.}
The deep$\leftrightarrow$maximal-shallow correspondence (theorem
\textsf{FaithfulSDlem},
Fig.~\ref{fig:MinFMLinHOL_faithfulness_locale}, lines~11--12) is
derived by induction on~$\varphi$ with
\texttt{auto simp:\,DefS\,DefD}; the global form \textsf{FaithfulSD}
(line~13) follows from \texttt{ValD\_def} and \texttt{ValS\_def}.
Similarly, the deep$\leftrightarrow$minimal-shallow correspondence is first shown in the
pointwise formulation \textsf{FaithfulMDlem}
(Fig.~\ref{fig:MinFMLinHOL_faithfulness_locale}, lines~17--18). The proof is a single \texttt{induct} on
$\varphi$ with rule \texttt{QInduct} (arbitrary in~$w$), with
\texttt{simp add:\,DefD\,DefM} discharging the propositional and modal
cases, and a final \texttt{blast} dispatching the existential case. The global form then follows from \texttt{ValM\_def}.

\paragraph{Lifted faithfulness.}
Inside \texttt{MinS\_ES}, the additional fact
$\mathcal{N}\_valid\_ES$ replaces \texttt{Range~gg} by~$DD$ in
\textsf{FaithfulMDlem}, yielding theorem \textsf{FaithfulMD\_ES}
(Fig.~\ref{fig:MinFMLinHOL_faithfulness_locale}, lines~31--32):
deep$\leftrightarrow$minimal-shallow faithfulness against the full
domain~$DD$ and universe~$WW'$ (validity still ranging over the worlds
of~$WW$), even when $DD$ and $WW'$ are uncountable.  The
specialisation to the universal domains, \texttt{MinS\_ES\_Univ},
yields the same correspondences against
$\langle\texttt{Univ},RR,II,\texttt{Univ}\rangle$.

\paragraph{Global faithfulness across all locale interpretations.}
The locale form admits a strong global theorem,
\textsf{FaithfulMS\_all}: a deep validity is precisely a validity
that holds against \emph{every} concrete model, and quantifying over
all locale interpretations of \texttt{MinS} captures exactly that
notion (Fig.~\ref{fig:MinFMLinHOL_faithfulness_locale},
lines~39--50).  The proof uses the \texttt{MinS\_specific} locale:
the right-to-left direction follows by unfolding deep validity and
applying \textsf{FaithfulMD}; the left-to-right direction unfolds
deep validity at an arbitrary model
$\langle WW,RR,II,DD\rangle,g_0,w$, instantiates
\texttt{MinS\_specific} for that model and the target formula, applies
\texttt{valid\_specific} to transfer truth to a constructed
surjective~$gg$, and invokes \textsf{FaithfulMD}.  In Isabelle this
is a short Isar proof of a dozen lines.  The operational meaning
is that \emph{any} statement provable inside the locale---i.e.\ from
purely minimal-shallow infrastructure---transfers directly to deep
validity.  We illustrate this in \S\ref{sec:exp-locale}.

\begin{figure}[htbp]
  \centering
  \colorbox{gray!30}{\includegraphics[width=.95\columnwidth]{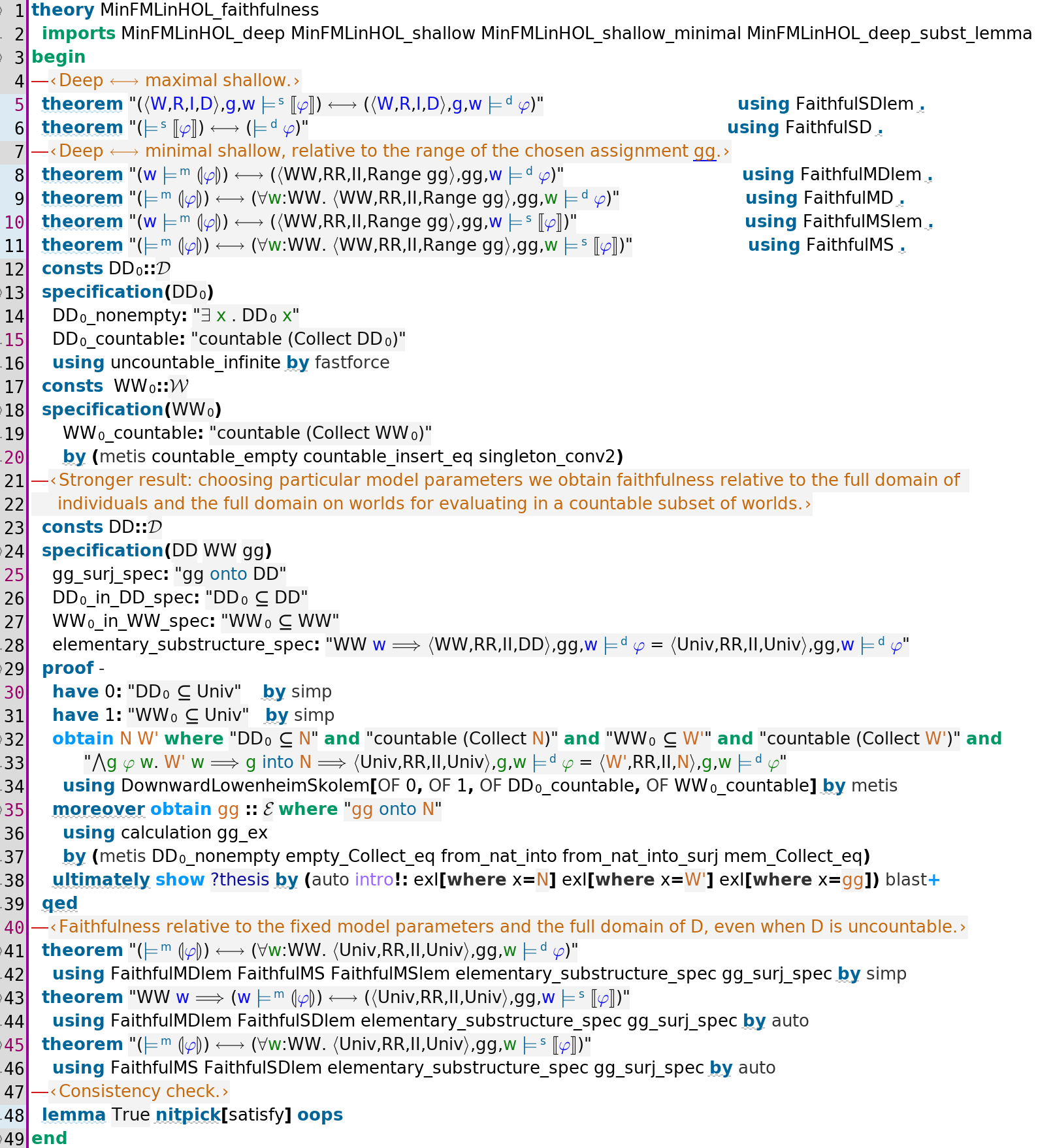}}
  \caption{Global (constants-level) re-issue of the faithfulness
  theorems (theory \texttt{MinFMLinHOL\_faithfulness.thy}): the
  deep$\leftrightarrow$maximal-shallow and
  deep$\leftrightarrow$minimal-shallow theorems relative to
  \texttt{Range gg} (lines~5--11); \texttt{specification} of a
  countable initial domain~$DD_0$ and world set~$WW_0$ (lines~12--20);
  the elementary sub-model $DD,WW,gg$ of the universal model obtained
  from \textsf{DownwardLowenheimSkolem}
  (\texttt{elementary\_substructure\_spec}, lines~24--39); and the
  faithfulness theorems re-issued against the universal domains
  (lines~41--46).
  \label{fig:MinFMLinHOL_faithfulness}}
\end{figure}

\paragraph{Global instance.}
For convenience, the theory \texttt{MinFMLinHOL\_faithfulness.thy}
(Figure~\ref{fig:MinFMLinHOL_faithfulness}) also re-issues the
faithfulness theorems at the constants level.  Using
\texttt{specification} it fixes a nonempty countable initial
domain~$DD_0$ and a countable initial world set~$WW_0$ (lines~12--20),
and then---directly from \textsf{DownwardLowenheimSkolem} and
\texttt{gg\_ex}---a domain~$DD$, world universe~$WW$ and surjective
assignment~$gg$ that form an elementary sub-model of the universal
model $\langle\texttt{Univ},RR,II,\texttt{Univ}\rangle$
(\texttt{elementary\_substructure\_spec}, lines~24--39).  The
locale-level faithfulness theorems are then re-issued at the global
level against the universal domains (lines~41--46).

\section{Experiments}\label{sec:exp}

We illustrate the development with the test file
\texttt{MinFMLinHOL\_experiments.thy}.  The
propositional tautologies (lines~7--12) and the K~axiom
(lines~14--16) are dispatched by \texttt{auto} after unfolding
\texttt{DefD}, \texttt{DefS}, or \texttt{DefM}; the necessitation rule
(lines~18--20) is similarly automatic.  Throughout
\texttt{MinFMLinHOL\_experiments.thy} every example theorem is stated
and proved in all three embedding variants (with the $d$, $s$, and~$m$
superscripts), to demonstrate that the alignment between the
embeddings is operational.

\paragraph{Barcan and converse Barcan.}
Both BF and CBF are valid in our constant-domain semantics; following
the d/s/m discipline, each is proved in the three embedding variants
with the uniform tactic
\texttt{unfolding DefD/DefS/DefM by auto}
(lines~26--31).
In a varying-domain semantics, neither BF nor CBF would be
HOL-derivable without further side conditions.  The
$\Diamond/\exists$ versions (lines~33--38) are equally
automatic, illustrating that under constant domain $\Diamond\exists x$
and $\exists x\Diamond$ coincide.

\paragraph{Disproofs by \texttt{nitpick}.}
\texttt{nitpick} reliably finds finite countermodels for invalid
schemata (lines~40--48).
The schema $\varphi\supset\Box\varphi$ (modal collapse, lines~44--45)
fails by a 2-world model in which $P$ holds at the start world but
not at its $R$-successor.  The schema $\Box\varphi\supset\varphi$
(lines~47--48) fails on a model with no reflexive edges,
illustrating the correspondence with the T~axiom.  Several such
disproofs use the auxiliary full-domain relation $\models^{d'}$,
marked ``Note:\ needs full domain~$D$''.

\subsection{Locale-based transfer in action}\label{sec:exp-locale}

\begin{figure}[htbp]
  \centering
  \colorbox{gray!30}{\includegraphics[width=.95\columnwidth]{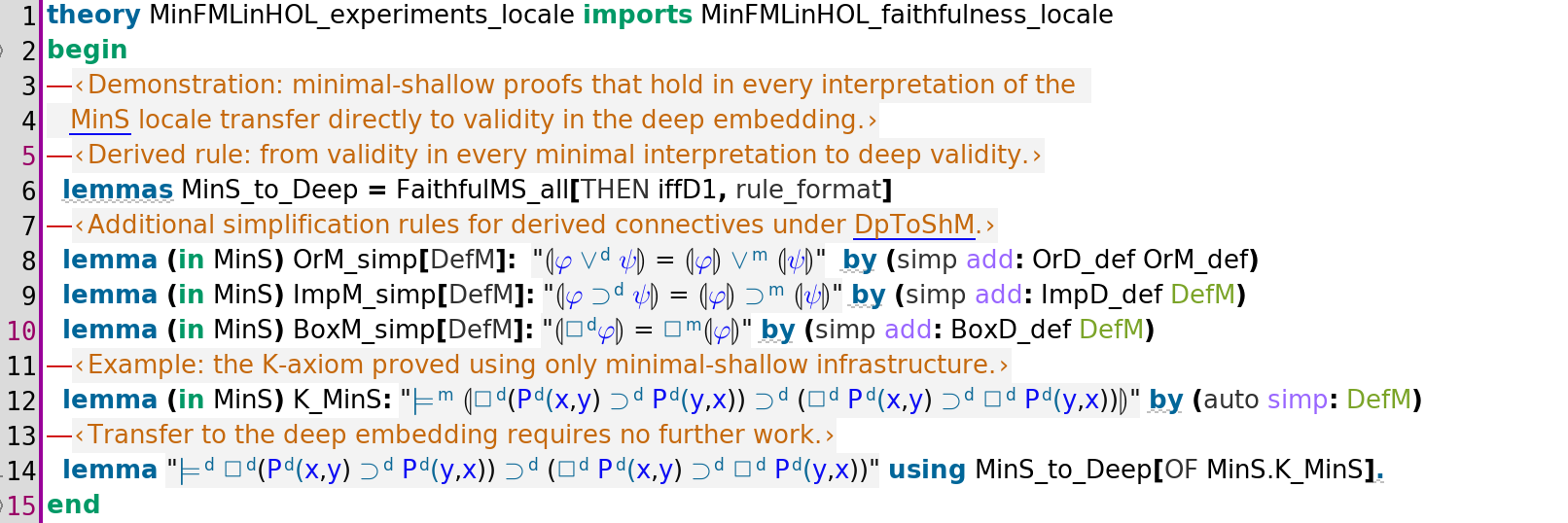}}
  \caption{Locale-based transfer to the deep embedding (theory
  \texttt{MinFMLinHOL\_experiments\_locale.thy}): the directional
  reading \texttt{MinS\_to\_Deep} of $\textsf{FaithfulMS\_all}$
  (line~6), three cosmetic locale-level rewrite rules
  \texttt{OrM\_simp}, \texttt{ImpM\_simp} and \texttt{BoxM\_simp} (lines~8--10), the
  K-axiom proved inside the locale (line~12), and its one-line
  transfer to deep validity (line~14).
  \label{fig:MinFMLinHOL_experiments_locale}}
\end{figure}

A short demonstration of \textsf{FaithfulMS\_all} is given in
\texttt{MinFMLinHOL\_experiments\_locale.thy}
(Figure~\ref{fig:MinFMLinHOL_experiments_locale}).  The K-axiom is
first proved purely from minimal-shallow definitions, inside the
locale (lemma \texttt{K\_MinS}, line~12), and transfer to deep
validity is then a single line (line~14), where
\texttt{MinS\_to\_Deep} (line~6) is the directional reading of
\textsf{FaithfulMS\_all}.  This is the operational pay-off of the
locale design: the user proves once in the simplest setting, and the
strong faithfulness theorem ports the result to the syntactic deep
embedding.

\paragraph{Auxiliary locale-level rewrites.}
The mapping \texttt{DpToShM} is defined only on the primitive cases
of the deep embedding.  Defined connectives such as $\supset^d$,
$\vee^d$ and $\Box^d$ are unfolded at the top of the deep embedding before
\texttt{DpToShM} is applied; for the experiments we add three
locale-level rewrite rules
(Fig.~\ref{fig:MinFMLinHOL_experiments_locale}, lines~8--10) to the
\texttt{DefM} bag, namely \texttt{OrM\_simp}, \texttt{ImpM\_simp} and \texttt{BoxM\_simp}.
These rewrites are entirely cosmetic but smooth the locale-level
proofs above.

\section{Related work and discussion}\label{sec:related}

The shallow-embedding approach to FML in HOL has a long history.
Benzm\"uller and Paulson~\cite{J21,J23} introduced a minimal shallow
embedding of higher-order modal logic in HOL targeting computational
metaphysics; that work, however, did not maintain a corresponding
deep embedding, and its faithfulness was not mechanised within HOL.
Recent work by Benzm\"uller and
Reiche~\cite{J58} on public announcement
logic shows the necessity of explicit dependencies on
evaluation domains in a heavyweight shallow embedding; our maximal
shallow embedding can be seen as a propositional skeleton of that
construction.  Other deep embeddings of FML or related logics in
proof assistants typically target metalogical results (e.g.\ 
completeness, cut-elimination) rather than mutual faithfulness with
shallow embeddings, and so are not geared towards the same
automated proof-and-disproof dialogue that is our focus.

\paragraph{First-order constant-domain mechanisations.}
Closest in subject matter is Kirst and
Shillito~\cite{KirstShillito2025}, who give a Coq-verified completeness
proof for first-order \emph{bi-intuitionistic} logic with constant
domains. Their aim is complementary to ours: a \emph{syntactic}
metatheorem (completeness of a calculus) for an \emph{intuitionistic}
object logic, versus our \emph{semantic} bridge (machine-checked
deep$\leftrightarrow$shallow faithfulness with automated proof and
\texttt{nitpick} disproof) for a \emph{classical} modal one. Where
they preserve constant domains through a completeness construction, we
must instead reconcile the deep embedding's quantification over an
arbitrary domain with the countable assignment of the minimal-shallow
locale --- precisely what the L\"owenheim--Skolem step of
\S\ref{sec:elem-subst} does.

\paragraph{Novelty and outlook.}
Kripke-semantics mechanisations are by now common, and we claim none
as such. New, to our knowledge, is their \emph{combination} in one
first-order modal HOL theory: a deep and two shallow embeddings kept
in \emph{machine-checked mutual faithfulness}, the locale-based global
theorem \textsf{FaithfulMS\_all}, and a constant-domain downward
L\"owenheim--Skolem theorem deployed to defuse the surjectivity
obstacle against uncountable domains. For the LogiKEy methodology~\cite{J48} this matters concretely: it
supplies, for the quantified modal case, both the syntactic deep
embedding hitherto lacking and a machine-checked guarantee that
shallow-level automation stays faithful to it.

\paragraph{Locales for shallow embeddings.}
Locales~\cite{ballarin2014} have been used widely in Isabelle to
parametrise algebraic structures, but their use to package shallow
embeddings of object logics, with a global faithfulness theorem
quantifying over all interpretations, appears to be new.  Each
locale parameter corresponds to a degree of semantic dependency that
the user is free to fix or to leave open.  The strong global
faithfulness theorem \textsf{FaithfulMS\_all} is the technical
pay-off and is, to our knowledge, novel.

\paragraph{The L\"owenheim--Skolem theorem in HOL.}
Within Isabelle, formalisations of the L\"owenheim--Skolem theorem
exist for first-order logic (e.g.~\cite{From2022}), but to our
knowledge none has previously
been formulated in the constant-domain Kripke-semantics setting of
FML, nor deployed to bridge a deep and a minimal-shallow embedding.
The mechanisation in \S\ref{sec:elem-subst} is fairly self-contained
($\sim 350$ lines including all auxiliary lemmata) and could be
reused for other shallow embeddings of quantified non-classical
logics that face the same surjectivity obstacle.

\paragraph{Availability.}
The full Isabelle/HOL development is bundled with this arXiv version as
\emph{ancillary files} (see the ``Ancillary files'' listing on the
abstract page) and consists of eleven theory files: preliminaries,
deep, deep-substitution-lemma, maximal-shallow, minimal-shallow locale
and global instance, faithfulness locale and global instance,
elementary-substructure, experiments and locale-based experiments,
together with an Isabelle \texttt{ROOT} session file.  It loads cleanly
in Isabelle2025-2 and Isabelle2026~\cite{Isabelle}.

\paragraph{Future work.}
The natural next step is to extend the present constant-domain
treatment to varying-domain semantics, which requires replacing the
rigid $\mathcal{E}:=V\Rightarrow D$ by a world-indexed
variable assignment $\mathcal{E}:=V\Rightarrow w\Rightarrow D$ and adjusting
the existential cases of all three embeddings simultaneously.  The
infrastructure we have built --- bag-based simplifier rule sets,
pointwise faithfulness lemmata, the size-based induction principle
\texttt{QInduct}, and the elementary-substructure construction ---
carries over without conceptual modification; in particular the
elementary-substructure construction already iterates over both the
individual and the world dimension, so only the world-indexing of the
domain itself remains to be added.  A second direction is to
generalise the modality to multimodal accessibility relations and
to inflationary/non-normal modalities, where the only change is in
the modal clause of every relevant function and the corresponding
clause of the substitution machinery.

\section{Conclusion}\label{sec:concl}

\textsf{MinFMLinHOL} demonstrates that the deep-and-shallow
LogiKEy methodology of~\cite{C98,JAIR-DeepShallow} scales smoothly to
logics with both binding and modality; it has meanwhile also been
applied to monadic second-order logic~\cite{MSOinHOL}.  The new ingredients are a
locale-based formulation of the minimal embedding, a corresponding
strong global faithfulness theorem, and a Tarski--Vaught-style
mechanisation of the (countable) downward L\"owenheim--Skolem
theorem that resolves the surjectivity problem against arbitrary
domains of individuals and worlds.  All faithfulness proofs are mechanised, with pointwise
lemmata and the substitution machinery carrying virtually all of the
work.

\section*{Declaration on Generative AI}
During the preparation of this work, the authors used generative AI assistants
(Anthropic's Claude family of models) in order to: draft initial prose for
several sections, suggest reformulations and shortenings, and
mechanise the bookkeeping of cross-references between the LaTeX
source and the Isabelle/HOL files.  The mathematical content, the
Isabelle/HOL theory development, and the specific design choices
(notably the locale-based packaging of the minimal embedding, the
formulation of \textsf{FaithfulMS\_all}, and the
elementary-substructure construction) are due to the authors.  All
text and all Isabelle/HOL proofs in the final manuscript and Archive
have been read, verified, and edited by the authors; the authors
take full responsibility for the content.

\bibliography{extra,chris,literature,bibliography}

\appendix

\section{Appendix}

The following figures show the rendered Isabelle/HOL source files
that supplement the body of the paper.  Line numbers refer to the
positions in the corresponding theory file.

\begin{figure}[!ht]
  \centering
  \colorbox{gray!30}{\includegraphics[width=.95\textwidth]{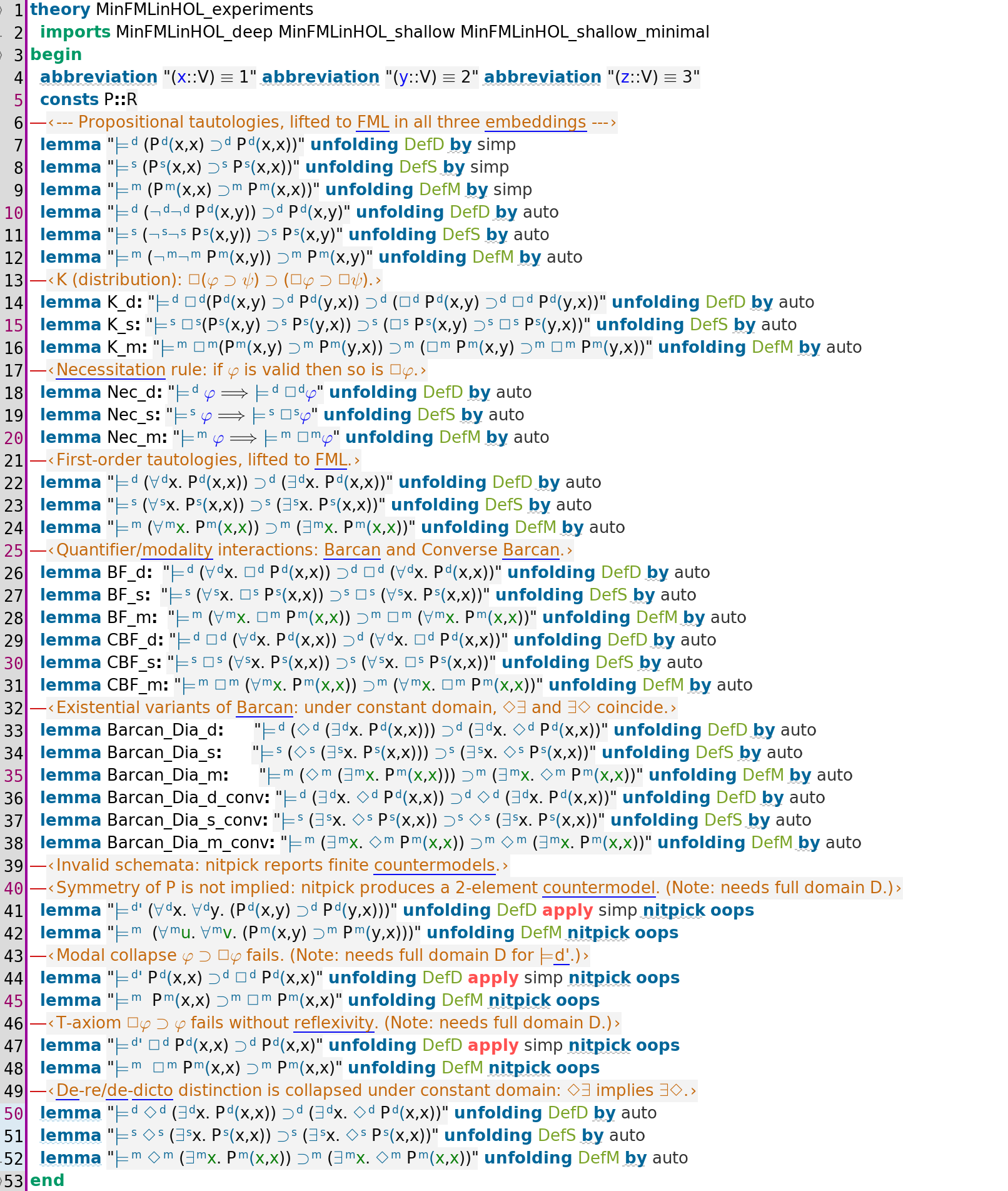}}
  \caption{Experiments in all three embeddings (theory
  \texttt{MinFMLinHOL\_experiments.thy}): propositional tautologies
  (lines~7--12), K-axiom (lines~14--16), necessitation (lines~18--20),
  first-order tautologies (lines~22--24), Barcan and converse Barcan
  formulas (lines~26--31), $\Diamond/\exists$ Barcan variants
  (lines~33--38), and disproofs by \texttt{nitpick} of (instances of) modal collapse,
  the T-axiom, and a non-symmetric atom (lines~40--48), followed by a
  restatement of the $\Diamond\exists\supset\exists\Diamond$ direction
  (lines~49--52).
  \label{fig:MinFMLinHOL_experiments}}
\end{figure}

\begin{figure}[!ht]
  \centering
  \colorbox{gray!30}{\includegraphics[width=.95\textwidth]{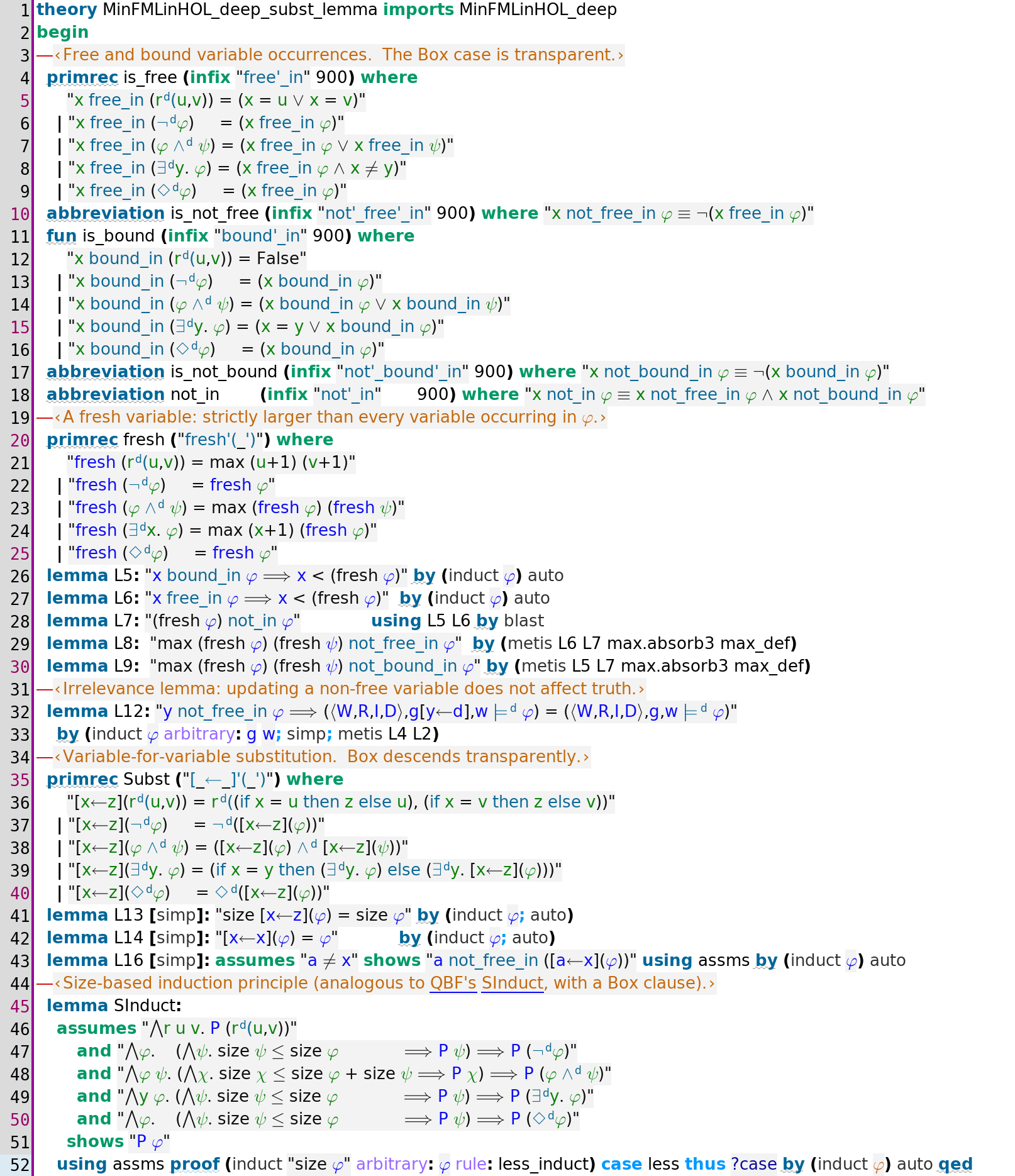}}
  \caption{Substitution machinery for the deep embedding (theory
  \texttt{MinFMLinHOL\_deep\_subst\_lemma.thy}, part~1): the
  free-/bound-variable predicates \texttt{is\_free} and
  \texttt{is\_bound} with their shorthands (lines~4--18), the
  fresh-variable function \texttt{fresh} (lines~20--25) and its
  consequence lemmata \texttt{L5}--\texttt{L9} (lines~26--30), the
  irrelevance lemma \texttt{L12} (lines~32--33), variable-for-variable
  substitution $[x{\leftarrow}z](\varphi)$ (lines~35--40) with lemmata
  \texttt{L13}, \texttt{L14} and \texttt{L16} (lines~41--43), and the
  size-based induction principle \texttt{SInduct} (lines~45--52).
  \label{fig:MinFMLinHOL_deep_subst_lemma_1}}
\end{figure}

\begin{figure}[!ht]
  \centering
  \colorbox{gray!30}{\includegraphics[width=.95\textwidth]{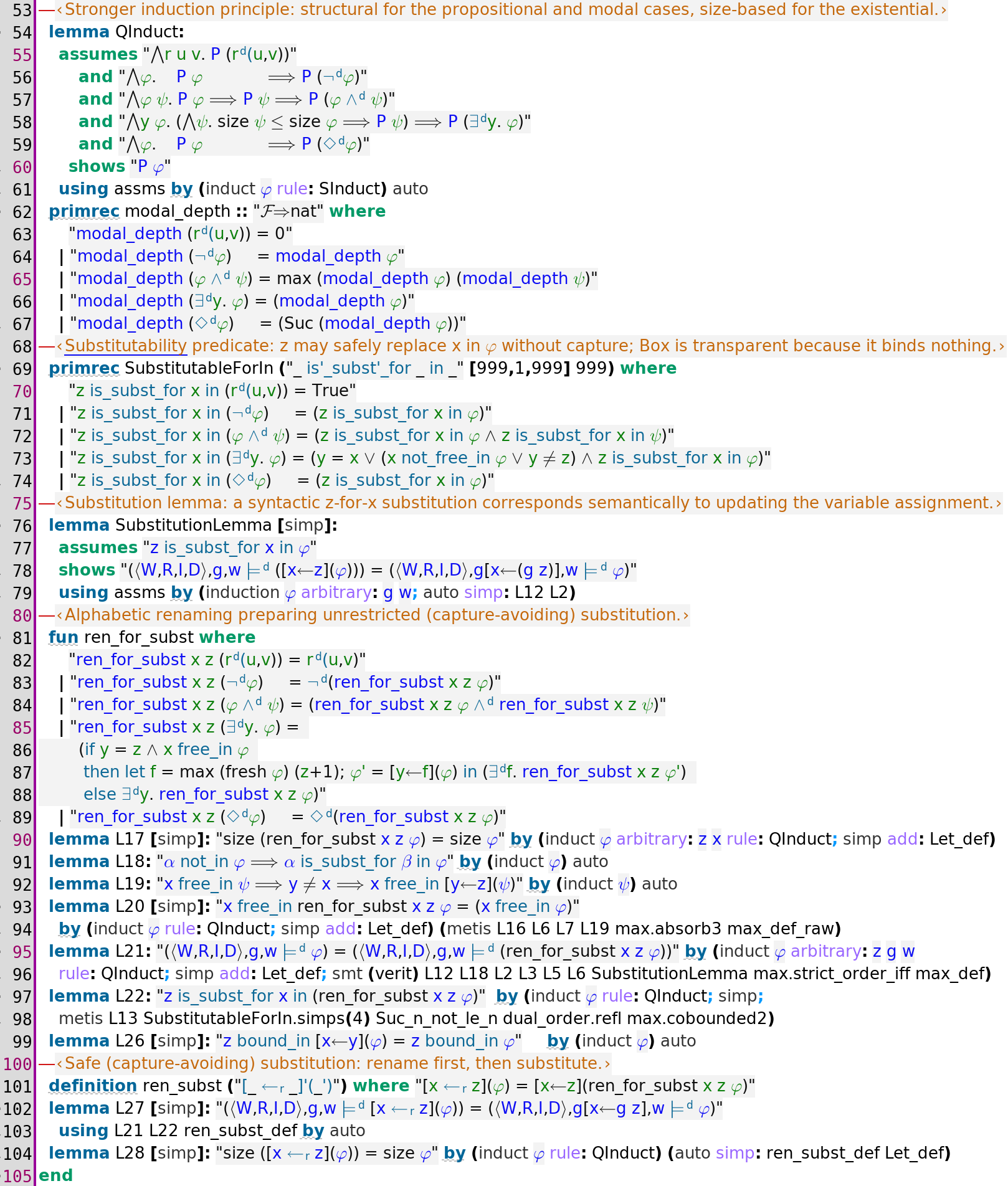}}
  \caption{Substitution machinery (theory
  \texttt{MinFMLinHOL\_deep\_subst\_lemma.thy}, part~2): the stronger
  induction principle \texttt{QInduct} (lines~54--61), the modal-depth
  function \texttt{modal\_depth} (lines~62--67), the substitutability
  predicate \texttt{is\_subst\_for} (lines~69--74), the substitution
  lemma \textsf{SubstitutionLemma} (lines~76--79), the
  alphabetic-renaming function \texttt{ren\_for\_subst} (lines~81--89)
  with lemmata \texttt{L17}--\texttt{L22} and \texttt{L26}
  (lines~90--99), and the safe-substitution operation
  \texttt{ren\_subst} written $[x{\leftarrow}_r z](\varphi)$ (line~101)
  with its semantic lemmata \texttt{L27} and \texttt{L28}
  (lines~102--104).
  \label{fig:MinFMLinHOL_deep_subst_lemma_2}}
\end{figure}

\begin{figure}[!ht]
  \centering
  \colorbox{gray!30}{\includegraphics[width=.95\textwidth]{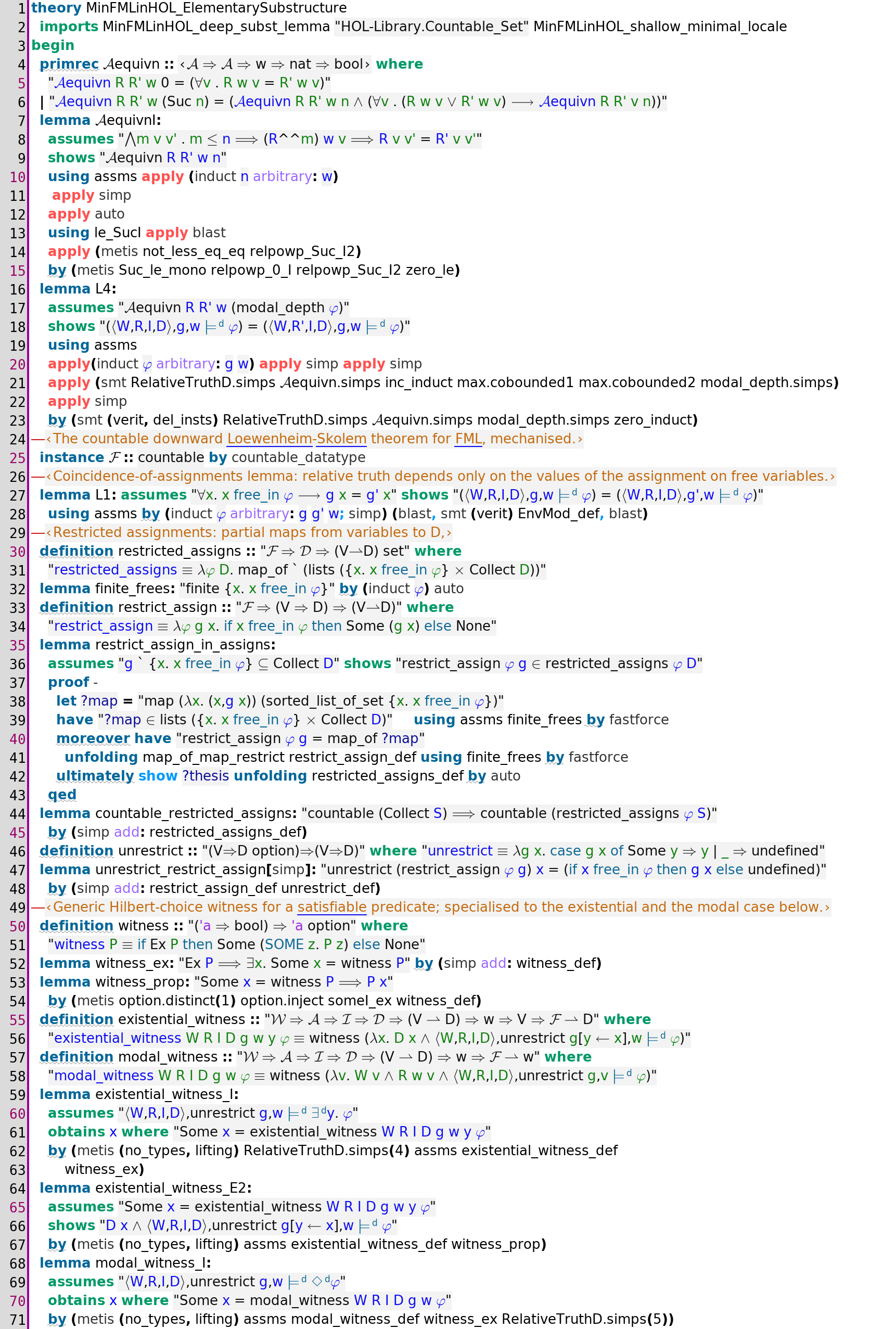}}
  \caption{The countable downward L\"owenheim--Skolem mechanisation
  (theory \texttt{MinFMLinHOL\_ElementarySubstructure.thy}, part~1):
  the $n$-step relation-equivalence predicate and lemma \texttt{L4},
  showing that deep truth depends on the accessibility relation only
  up to modal depth (lines~4--23); the coincidence-of-assignments
  lemma \texttt{L1} (lines~27--28); the partial-map machinery
  \texttt{restricted\_assigns}, \texttt{restrict\_assign} and
  \texttt{unrestrict} with their countability lemma (lines~30--48);
  and the generic Hilbert-choice \texttt{witness} operator
  (lines~50--54) together with its specialisations
  \texttt{existential\_witness} and \texttt{modal\_witness} and their
  accessor lemmata (lines~55--71).
  \label{fig:MinFMLinHOL_ElementarySubstructure_1}}
\end{figure}

\begin{figure}[!ht]
  \centering
  \colorbox{gray!30}{\includegraphics[width=.95\textwidth]{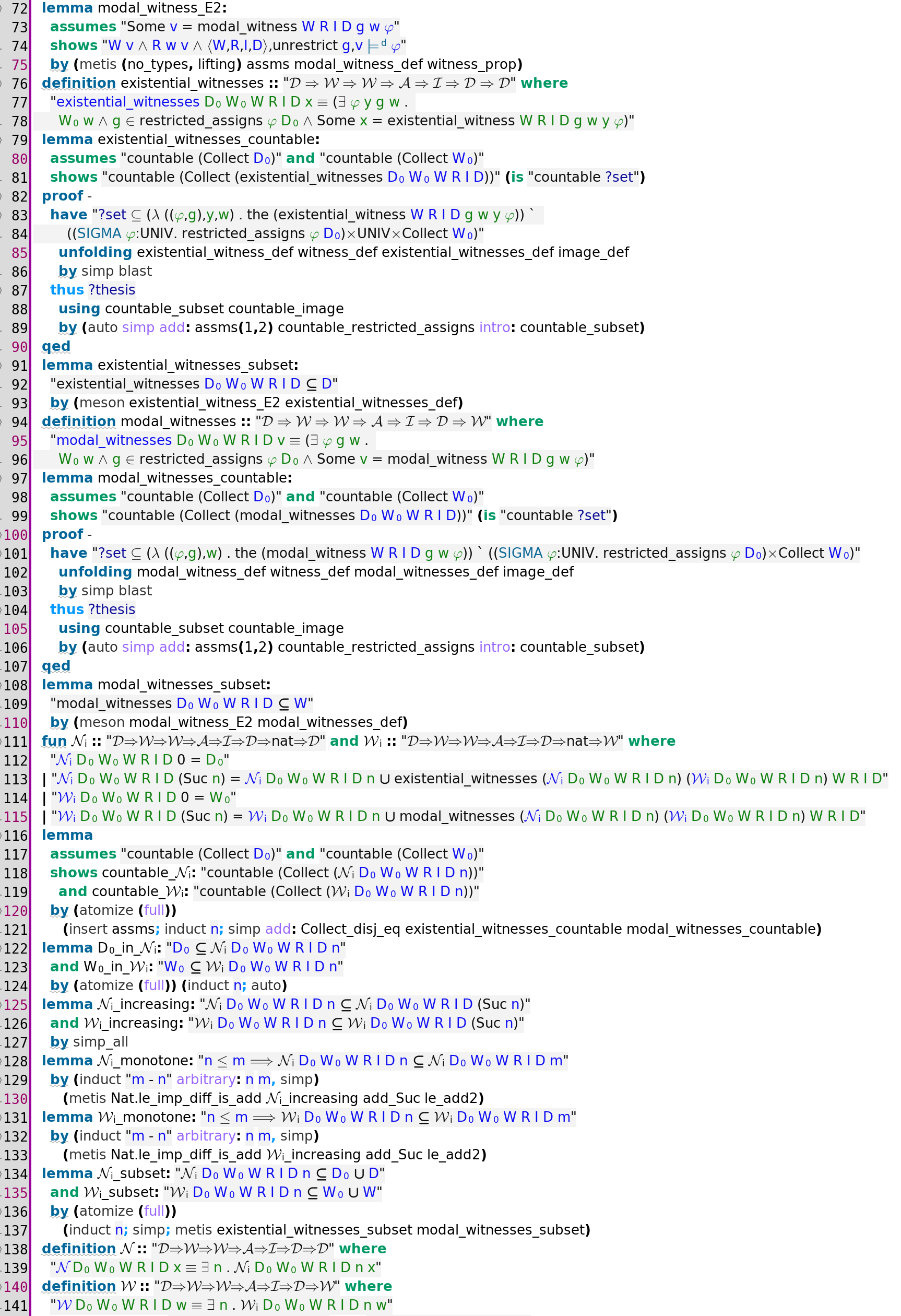}}
  \caption{The countable downward L\"owenheim--Skolem mechanisation
  (theory \texttt{MinFMLinHOL\_ElementarySubstructure.thy}, part~2):
  the witness-collection definitions \texttt{existential\_witnesses}
  and \texttt{modal\_witnesses} with their countability and subset
  lemmata (lines~76--110); the mutually recursive stages
  $\mathcal{N}_i$ and $\mathcal{W}_i$ (lines~111--115) with their
  joint countability, containment, monotonicity and subset lemmata
  (lines~116--137); and the limits $\mathcal{N}$ and $\mathcal{W}$
  (lines~138--141).
  \label{fig:MinFMLinHOL_ElementarySubstructure_2}}
\end{figure}

\begin{figure}[!ht]
  \centering
  \colorbox{gray!30}{\includegraphics[width=.95\textwidth]{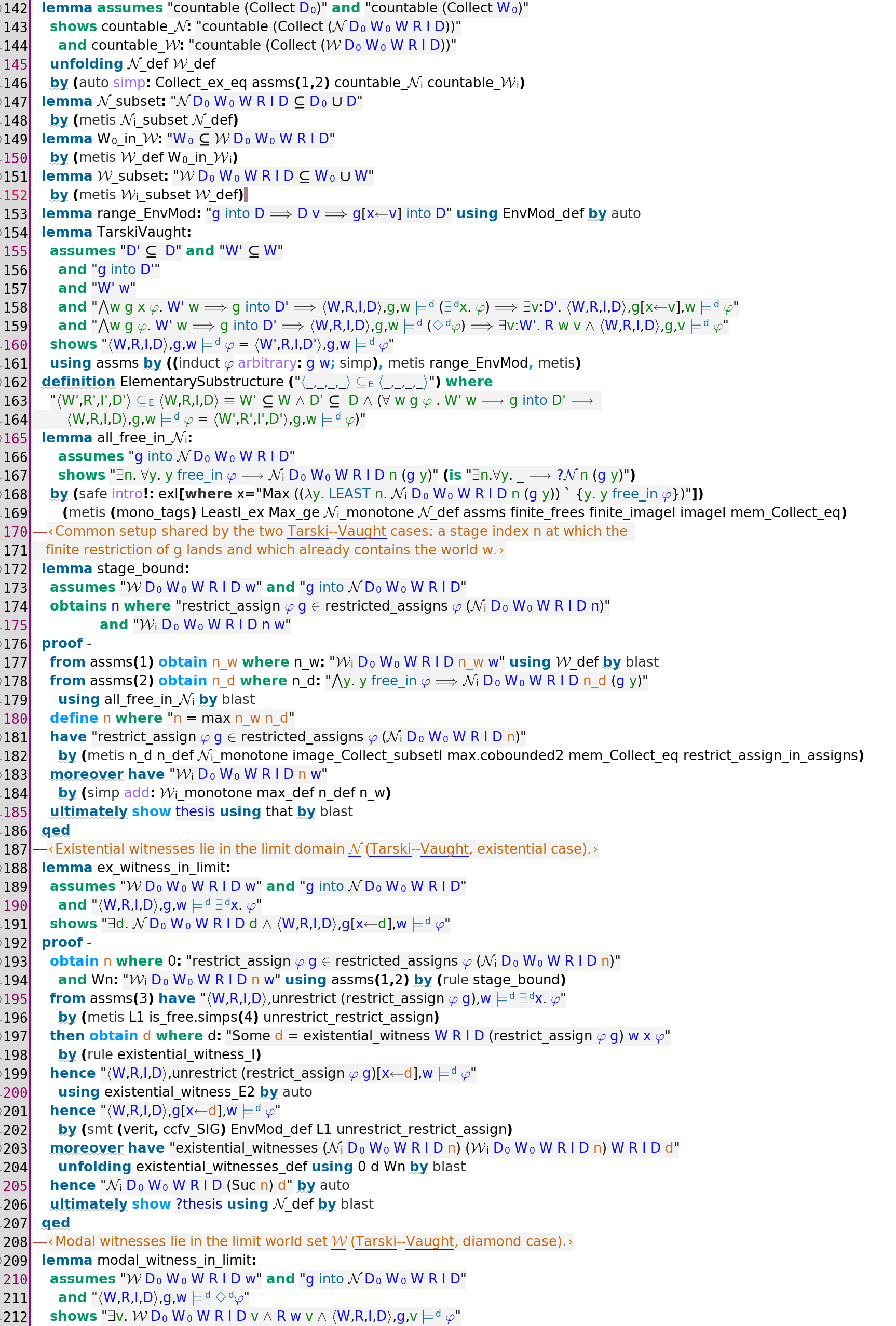}}
  \caption{The countable downward L\"owenheim--Skolem mechanisation
  (theory \texttt{MinFMLinHOL\_ElementarySubstructure.thy}, part~3):
  the countability, containment and subset lemmata for the limits
  $\mathcal{N}$ and $\mathcal{W}$ (lines~142--152); the
  Tarski--Vaught test \texttt{TarskiVaught} (lines~154--161); the
  elementary-substructure relation $\subseteq_E$
  (\texttt{ElementarySubstructure}, lines~162--164); the shared-setup
  lemma \texttt{stage\_bound} (lines~172--186); and the existential
  witness-in-limit lemma \texttt{ex\_witness\_in\_limit}
  (lines~188--207), with \texttt{modal\_witness\_in\_limit} beginning
  at line~209.
  \label{fig:MinFMLinHOL_ElementarySubstructure_3}}
\end{figure}

\begin{figure}[!ht]
  \centering
  \colorbox{gray!30}{\includegraphics[width=.95\textwidth]{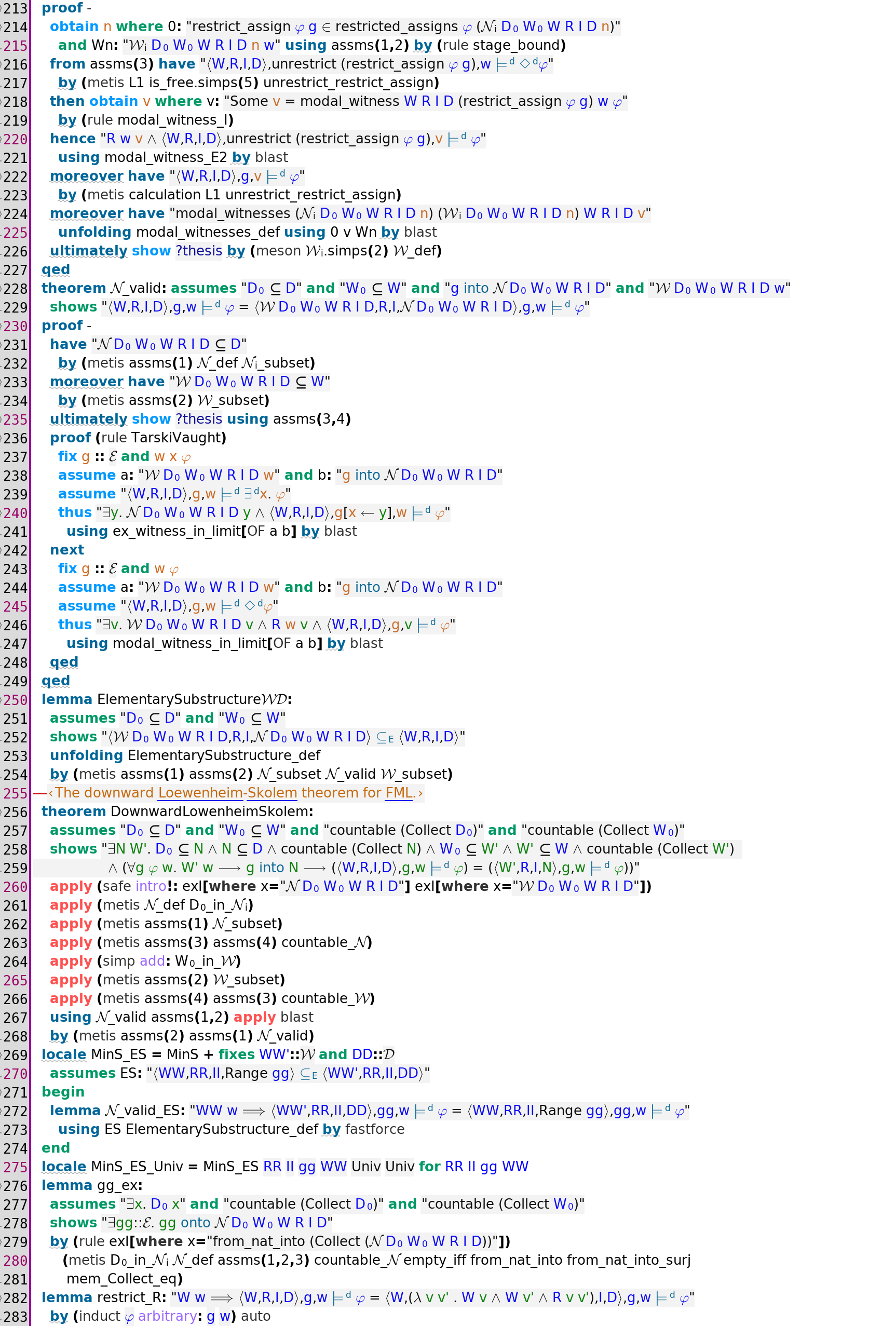}}
  \caption{The countable downward L\"owenheim--Skolem mechanisation
  (theory \texttt{MinFMLinHOL\_ElementarySubstructure.thy}, part~4):
  the modal witness-in-limit lemma \texttt{modal\_witness\_in\_limit}
  (through line~227); the pointwise elementary-substructure theorem
  $\mathcal{N}\_valid$ (lines~228--249) and its packaging
  \texttt{ElementarySubstructure}$\mathcal{WD}$ (lines~250--254); the
  headline theorem \textsf{DownwardLowenheimSkolem} (lines~256--268);
  the locales \texttt{MinS\_ES} and \texttt{MinS\_ES\_Univ} with lemma
  $\mathcal{N}\_valid\_ES$ (lines~269--275); and the auxiliary
  lemmata \texttt{gg\_ex} and \texttt{restrict\_R} (lines~276--283).
  \label{fig:MinFMLinHOL_ElementarySubstructure_4}}
\end{figure}

\begin{figure}[!ht]
  \centering
  \colorbox{gray!30}{\includegraphics[width=.95\textwidth]{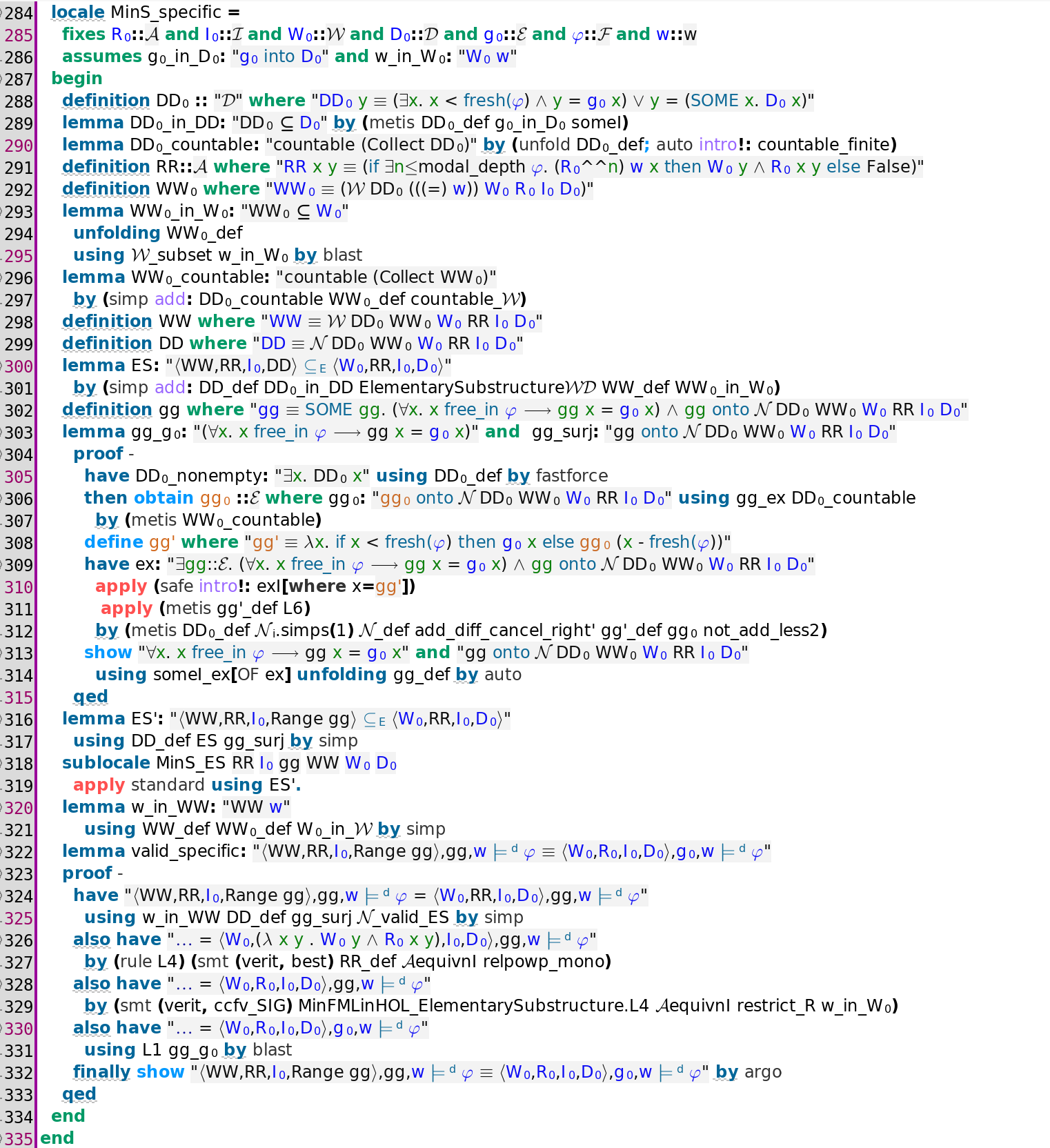}}
  \caption{The countable downward L\"owenheim--Skolem mechanisation
  (theory \texttt{MinFMLinHOL\_ElementarySubstructure.thy}, part~5):
  the locale \texttt{MinS\_specific}, which packages the particular
  construction powering global faithfulness --- a seed sub-domain
  \texttt{DD}$_0$, a bounded accessibility relation \texttt{RR}, the
  world and domain carriers \texttt{WW}/\texttt{DD}, and a surjective
  assignment \texttt{gg} agreeing with a given $g_0$ on the free
  variables of the target formula (lines~284--319) --- culminating in
  the transfer lemma \texttt{valid\_specific} (lines~322--333).
  \label{fig:MinFMLinHOL_ElementarySubstructure_5}}
\end{figure}

\end{document}